\pdfoutput=1

\documentclass[11pt]{article}

\usepackage[final]{acl}

\usepackage{times}
\usepackage{latexsym}

\usepackage[T1]{fontenc}

\usepackage[utf8]{inputenc}

\usepackage{microtype}

\usepackage{graphicx} 

\usepackage{hyperref}

\usepackage{threeparttable}
\usepackage{tabularx}
\usepackage{multirow}

\author{Martin Hyben, Sebastian Kula, Ivan Srba, Robert Moro, Jakub Simko \\
   Kempelen Institute of Intelligent Technologies \\
   Bottova 7939/2A,\\ 811 09 Bratislava \\
  \texttt{\small  \{martin.hyben, sebastian.kula, ivan.srba, robert.moro, jakub.simko\}}\small @kinit.sk \\}

\title{Multilingual and Multi-topical Benchmark of Fine-tuned Language models and Large Language Models for Check-Worthy Claim Detection}

\date{October 2023}

\begin{document}

\maketitle

\begin{abstract}
    This study compares the performance of (1) fine-tuned language models and (2) large language models on the task of check-worthy claim detection. For the purpose of the comparison we composed a multilingual and multi-topical dataset comprising texts of various sources and styles. Building on this, we performed a benchmark analysis to determine the most general multilingual and multi-topical claim detector.
    We chose three state-of-the-art models in the check-worthy claim detection task and fine-tuned them. Furthermore, we selected four state-of-the-art large language models without any fine-tuning. We made modifications to the models to adapt them for multilingual settings and through extensive experimentation and evaluation, we assessed the performance of all the models in terms of accuracy, recall, and F1-score in in-domain and cross-domain scenarios. Our results demonstrate that despite the technological progress in the area of natural language processing, the  models fine-tuned for the task of check-worthy claim detection still outperform the zero-shot approaches in cross-domain settings.
\end{abstract}

\section{Introduction}
In an ongoing information war that is waged in the online space, it is crucial to be able to filter out the important information from the massive amount of data published every day. \emph{Check-worthy claim detection} plays a crucial role in manual, automated and also fully automatic fact-checking and information verification, facilitating the identification of statements that require further investigation.

As a check-worthy claim is considered such a factual statement which can cause a potential harm (to a society, to a health of individuals, etc.) and thus is worth to be fact-checked. In Table~\ref{tab:check-worthy-ex} we present an example of a check-worthy and non-check-worthy claim. The first claim is check-worthy due to its potential impact on public health and safety, warranting a fact-check to ensure accurate information for the public. The second claim is non-check-worthy due to extensive debunking and overwhelming evidence, making fact-checking redundant and of minimal value.

\footnotetext[1]{We will share our source code, including the integration mechanism of the benchmarking dataset and trained models on the Github and on the Zenodo upon the paper acceptance.}
\setcounter{footnote}{1}

\begin{table}
    \centering
    \small
    \setlength\tabcolsep{2pt}
    \begin{tabular}{p{5cm}c}
        \hline
        \textbf{Claim text} & \textbf{Check-worthy} \\
        \hline
        "A new study suggests that drinking bleach can effectively cure COVID-19." & True\\
        "We have been to the Moon." & False\\
        \hline
    \end{tabular}
    \caption{Example of a check-worthy and non-check-worthy claim.}
    \label{tab:check-worthy-ex}
\end{table}

Since the misinformation come in many topics, languages and writing styles, the problem of check-worthy claim detection applies not only to the news articles with well polished and structured content, but also to social media discussions with often unstructured and noisy content, as well as to multiple languages or topical domains. Nevertheless, the current state-of-the-art models are mostly specialized to narrow domains, limited to small subsets of topics or languages, or exclusively to English~\cite{stammbach2023environmental,reddy2022newsclaims}. While current research is constrained by limitations in existing works, this paper seeks to conduct a more comprehensive comparison with the goal of identifying the most general model. Consequently, there is a lack of an appropriate benchmark dataset or study that offers explicit insights into the comparative performance of these models and their ability to generalize across diverse out-of-domain scenarios.

In this study\footnotemark[1], we therefore for the first time surveyed the existing datasets and models for the task of Check-worthy claim detection. As the state-of-the-art is focused on narrow sub-problems limited in topical, linguistic and formal variability, we constructed a \emph{general claim dataset} that would fit the task more broadly. It is composed of multiple existing datasets with emphasis on the variability of topics, languages and writing styles.
Utilizing the obtained general claim dataset, we provide a novel comparison of the state-of-the-art large language models (LLMs), fine-tuned specifically for the task of Check-worthy claim detection.

In addition, with the emergence of LLMs there was a widespread belief that such powerful models would be able to perform any task in the NLP domain without the requirement for fine-tuning, and to perform it better than a human or models created so far, which are based on much smaller architectures than LLMs~\cite{DBLP:journals/corr/abs-2302-08091, How_ChatGPT}. They have already been used for zero-shot identification of factual claims~\cite{li2023selfchecker} as well as in~\cite{CheckThatLab-GPTvsBERT} where authors conducted a comparison of ChatGPT and BERT model fine-tuned on CheckThat! Lab's check-worthy claims extraction task. These studies however focused exclusively on limited scenarios and did not covered the most recent LLMs like GPT-4, which are expected to perform better as the previous generation of the LLMs.

In this study, we therefore compared the performance of fine-tuned language models with the most recent LLMs (like GPT-4) applied in zero-shot setting. Such so-far-missing comparison provides insights on how the most recent advancement compare to the state-of-the-art of a check-worthy claims detection task.

Our main contributions are following:
\begin{enumerate}
    \item A comprehensive benchmark of state-of-the-art models for the task of check-worthy claim detection in multilingual and mutli-topic setting. 
    \item A composite check-worthy claim dataset with content spanning various sources (social-media, news articles, reports, blogs), of various nature: noisy, semi-noisy, structured, and also various topics (Environment, Health, Politics, Science, Sports, Entertainment) and languages (Arabic, Bulgarian, Dutch, English, Spanish, Turkish, Slovak, Czech, Polish and Hungarian).
    \item A comparison between fine-tuned language models and zero-shot LLMs showing that, at this point, even smaller fine-tuned language models outperform LLMs with unaltered configurations. 
\end{enumerate}

\section{Related work}
\label{sec:related_work}
Large Language models (LLMs) initially introduced in ~\cite{brown2020language} are still relatively new, however, in a recent past many such models gained a lot of attention (e.g., ChatGPT, Bard, BingAI, or GPT4.0). Yet, they have already seen applications in the area of disinformation tackling.
\citet{journals/corr/abs-2306-17176} makes a comparative analysis of four models (chatGPT3.5, Bard, BingAI, GPT4.0) based on LLM architecture used in news fact-checking. Systematic evaluation to measure ChatGPT’s fact-checking performance using a zero-shot classification approach was carried out in~\cite{hoes_altay_bermeo_2023}. \citet{journals/corr/abs-2303-08652} address the issue of verifying facts by automatically generating search queries in order to place them in the search engine and to retrieve the evidence. Search queries are formulated using the T5 LLM and factual statements, which are collected in the evidence dataset. Another work concerns the verification of the credibility of news media domains using the ChatGPT~\cite{journals/corr/abs-2304-00228}. \citet{journals/corr/abs-2305-04812} analyze the impact of false information on the behavior and responses generated by LLM, the ChatGPT and Alpaca-LLaMA models. \citet{vykopal2023disinformation} makes a comprehensive study of the disinformation capabilities of the current generation of LLMs to generate false news articles in English.

More works in the check-worthy claim detection exist featuring the methods based on fine-tuned language models. \citet{gupta-etal-2021-lesa} proposed a generalized claim detection system LESA based on the BERT architecture (used for semantic features extraction) and BiLSTM (used for linguistic features extraction). The limitation is that the employed datasets are only in English and that the noisy data used in the training of the models were dedicated solely to the subject of COVID-19 (semi-noisy and non-noisy data are on various topics). Nevertheless, the LESA model has the feature of being general due to its ability to detect claims in any online text, regardless of whether the source of the text is a social media, a blog or a news article. 

The subject of the work of~\citet{DBLP:conf/emnlp/SundriyalKPA022} was to automatically identify claim spans in Twitter posts. The authors used the DABERT adapter-based variation of RoBERTa for claim span identification. The main limitation of this work is that the training data come from a single source (Twitter). \citet{DBLP:journals/corr/abs-1809-08193} used a universal sentence approach to perform claim detection. \citet{DBLP:conf/emnlp/AlamSDSNMMADDAZ21} focus on the application of the fine-tuned transformer-based LLMs (BERT, RoBERTa, araBERT, BERTje, mBERT, XLM-RoBERTa) to detect posts that contain check-worthy claims and harmful claims for society. The used dataset is limited only to tweets about COVID-19. \citet{DBLP:journals/corr/abs-2209-00507} detect environmental claims using models based on transformers architecture, such as distilBERT, RoBERTa and ClimateBERT. This work is topically very narrow and the obtained models work only for the detection of claims in articles in English and related to environmental issues. \citet{DBLP:conf/emnlp/ReddyCWFCEPNHSJ22} apply the ClaimBuster tool for the claim detection task, however, the datasets used are only in English, refer only to the topic of COVID-19 and the source of news articles are only news media.
\citet{DBLP:conf/acl/GuptaWLX22} focus on the task of verifiable claim detection in dialogues. They proposed a new benchmark and dataset, and also conducted an analysis of three methods for verifiable claim detection: Lexical overlap, DNLI (Dialogue Natural Language Inference) and Lexical + DNLI. Similarly to the previous works, the experiments are conducted on English only data. 

Finally, \citet{DBLP:conf/clef/NakovBMAMCKZLSM22} provide a review of the methods and results obtained in the CLEF'22 CheckThat! Lab. One of the subtasks of this competition was to detect check-worthy claims in tweets. The datasets covered six languages and related mainly to the topic of COVID-19. Participants of the competition used mainly transformer-based fine-tuned language models. The top results, depending on the language of the dataset, were achieved by mT5 (AR, BG, NL, ES), AraBERT (AR), Roberta (EN), BERT (TR) and GPT-3 (EN). The limitations of these results are that the models, in most of the cases were restricted to a single language (only three models were trained on more than one language at once but were not tested in multilingual mode at once) and a single (COVID-19) topic.

\section{Datasets}

\paragraph{Existing datasets.}
The task of this study was to identify the most general claim detector model capable of detecting the claims across multiple topics, languages and writing styles. The selection criteria therefore, focused on topic diversity, multilingualism and writing style variability, ensuring the robustness of our findings. As identified in~\cite{DBLP:conf/clef/NakovBMAMCKZLSM22}, we considered four distinct categories of claims, namely Check-worthy claims, Harmful claims, Attention-worthy claims, and Verifiable factual claims. Given our research focus and objectives, we chose to prioritize the evaluation of models on the Check-worthy claims category. This decision was motivated by the fact that Check-worthy claims encompass all other categories as a power set, allowing us to assess model performance on a comprehensive and diverse set of claims. By concentrating on the Check-worthy claims, we aimed to evaluate the models' generalization capabilities on the most extensive range of claim types, ensuring a robust assessment of their effectiveness in handling various real-world social media and web content scenarios. We have identified multiple publicly available datasets that fit the aforementioned criteria. Their summary is presented in Table~\ref{tab:datasets-topics}; a comprehensive list together with their detailed description can be found in Appendix~\ref{sec:appendix-datasets}.

\paragraph{General claim dataset.}
We placed emphasis to ensure the variability of included topics, languages and writing styles. Based on these criteria, we selected five datasets from the list of publicly available datasets as described in Table~\ref{tab:datasets-topics} to be included in our \textit{General Claim dataset}.

\begin{table*}
    \centering
    \small
    \setlength\tabcolsep{2pt}
    \begin{threeparttable}
        \begin{tabularx}{\textwidth}{llXXll}
            \hline
            \textbf{Dataset} & \textbf{Samples} & \textbf{Topic} & \textbf{Languages} & \textbf{Sources} \\
            \hline
            CLEF'22 & 62,703 & Health, Politics & ar, bg, du, en, sp, tr & Twitter & ~\cite{DBLP:conf/clef/NakovBMAMCKZLSM22} \\
            CLEF'23 & 238,510 & Health, Politics & ar, en, sp & Twitter & ~\cite{clef-checkthat:2023:task1} \\
            MultiClaim & 200,000 & Environment, Health, \newline Politics, Science, Sports,  \newline Entertainment &  ar, bg, du, en, sp, tr, sk,  \newline cz, pl, hu  \newline (out of 39 languages) & Fact-checks & ~\cite{pikuliak2023multilingual} \\
            LESA & 379,348 & Health, Politics  & en & Twitter & ~\cite{gupta-etal-2021-lesa} \\
            EnvClaims & 29,400 & Environment & en & Reports\tnote{1} & ~\cite{stammbach2022dataset} \\
            \hline
            \textbf{General claim} & \textbf{420,738} & \\
            \hline
            NewsClaims & 1758 & Politics & en & News articles & ~\cite{reddy2022newsclaims} \\
            ClaimBuster & 23,533\tnote{3} & Politics & en & Presidental debates\tnote{2}  & ~\cite{10.14778/3137765.3137815} \\ 
ar, en, sp Twitter        \end{tabularx}
        \begin{tablenotes}
            \item[1] Sustainability reports, earning calls and annual reports of listed companies.
            \item[2] U.S. general election presidential debates.
            \item[3] Used only a subset of 970 samples for purpose of this benchmark.
        \end{tablenotes}
        \caption{A compilation of datasets used in this study. The General claim dataset is a combination of 5 selected datasets. Total number of samples includes balancing and filtering out the samples based on specified criteria.}
        \label{tab:datasets-topics}
    \end{threeparttable}
\end{table*}

To ensure the consistency and comparability of the samples, we carefully harmonized the datasets in terms of included languages and number of samples. Initially, we filtered out any empty or duplicate samples and aggregated samples from all 10 languages encompassed by our selected datasets. To ensure the uniformity, we created a histogram of word count for each dataset. To compensate the major differences in the word count for each dataset, we employed the abstractive text summarization using the Sumy TextRank summarizer ~\cite{mihalcea-tarau-2004-textrank} to distill each sample from both the CLEF'22 and CLEF'23 datasets into a single sentence, thereby simplifying the content. Based on the overlap of the histogram ranges after the compensation, we constrained the word count of each sample in all datasets to fall within the range of 4 to 30 words. Subsequently, we took measures to balance the resulting dataset, ensuring an equal distribution of samples. 

For uniformity assurance within the benchmark, we applied a standardized preprocessing procedure across all datasets included in our study. This preprocessing method,  conducted using the Pandas Python package version 1.5.3, involved the removal of special characters, such as emoticons and hashtags, to ensure that the text was devoid of non-linguistic elements. Additionally, we eliminated alphanumerical characters to maintain data integrity. 
To assess the influence of the preprocessing procedure on the final outcomes, we conducted a thorough two-fold evaluation: one with the preprocessing procedure applied, and the other without it. By contrasting these two approaches, we aimed to discern the optimal solution, shedding light on the significance of the preprocessing step in the final results. In our analysis of 210 conducted experiments, we found that 88 of them, or 42\%, demonstrated improvement when employing beforementioned preprocessing techniques, while the remaining 122, or 58\%, did not exhibit any enhancements.

\section{Fine-tuned language models}

The following models were used in the presented work: XLM-RoBERTa~\cite{DBLP:conf/acl/ConneauKGCWGGOZ20}, LESA~\cite{gupta-etal-2021-lesa} mDeBERTa ~\cite{DBLP:conf/iclr/HeGC23}, which were trained during experiments in order to create models capable of detecting check-worthy claims in the widest achievable range. These models were chosen as XLM-RoBERTa and mDeBERTa are multilingual and LESA was initially created as the model for detecting check-worthy claims in texts with a non-uniform subject and structure. The goal was to make the models for the detection of check-worthy claims universal and capable of detecting claims in texts on various topics and regardless of the type of text (noisy, semi-noisy, non-noisy). The detailed description of used models can be found in Appendix~\ref{sec:appendix-finetuned}.

We replicated all models according to the original papers and performed a comparison with original results. Our results closely matched the original findings. A comprehensive breakdown of the results can be located in the Appendix ~\ref{sec:appendix-in-dist}.

\section{Large Language Models}

The second group of methods applied are methods based on the concept of prompting and in-context learning, which uses LLMs in a few-shot or zero-shot way. In order to create a comprehensive overview of the current state of language models, we carefully selected four models that represent different aspects of the field. Alpaca-LoRA-7B was chosen as it was reported as a successful example of low-rank adaptation model, which performs comparable to the Stanford Alpaca model ~\cite{Alpaca-LoRA}. Mistral-7B represents the latest advancements in this area, and as was reported outperforms Llama 2 13B on all benchmarks ~\cite{jiang2023mistral}. Additionally, we included ChatGPT and GPT-4.0 as they are among the most widely used and highly regarded models in the community, and are often considered benchmarks for evaluating other language models. In all four of these models, the prompting mechanism was used, by introducing a question (instruction) whether the text in the input contain check-worthy claims. In this work the in-context learning mechanism was not applied, even though the models have capability to efficiently conduct the mechanism.

The ChatGPT is a transformer-based model developed using GPT-3.5 and GPT-4.0. The creator of ChatGPT, OpenAI, in order to facilitate the work of users, has created a dedicated APIs, which were applied in this presented work. Details of specific models used and their configuration is discussed in the Appendix \ref{sec:appendix-LLMs}

The LLaMA Alpaca-LoRA is a low rank adaptation (LoRA) of Stanford Alpaca~\cite{Alpaca-LoRA}, which is a reduced adaptation of LLaMA 7b. The LoRA mechanism is frequently used for LLM models and allows for greatly reducing of trainable parameters and thus creating lightweight versions of models~\cite{DBLP:journals/corr/abs-2303-18223, DBLP:conf/iclr/HuSWALWWC22}. In the task of detecting check-worthy claims, the API offered by the creators of the LLaMA Alpaca-LoRA was used.

Mistral-7B, as described in ~\cite{jiang2023mistral}, incorporates Grouped-Query Attention (GQA) ~\cite{ainslie2023gqa} and Sliding Window Attention (SWA) ~\cite{child2019generating, beltagy2020longformer}. GQA accelerates inference speed, reduces memory requirements during decoding, enabling higher batch sizes for increased throughput in real-time applications. SWA efficiently handles longer sequences at a lower computational cost, overcoming a common limitation in LLMs. The combined use of these attention mechanisms contributes to Mistral-7B's improved performance and efficiency, as highlighted in ~\cite{mukherjee2023orca}.

\paragraph{Prompting and prompt engineering techniques.}
In our experiments with LLMs, we maintained a consistent prompting technique to ensure uniformity and ease of interpretation. The primary prompt used was, \textit{``Does the input contain a check-worthy claim? Answer strictly in binary \textit{Yes} or \textit{No}.''} However, we also explored a range of alternative prompts to assess their impact on the models' performance and the interpretability of their output. 
Throughout these experiments, we meticulously analyzed the models' responses and adjusted the prompts as needed to align with the criteria for check-worthy claim detection. Our objective was not only to enhance the models' performance but also to streamline the interpretation of the LLM's output, enabling us to treat it as a binary class label (i.e., `Yes' for a check-worthy claim and `No' otherwise) for subsequent analysis and evaluation.

\section{Experimental methodology}
We divided the seven publicly available datasets into two groups: 1) a \textit{train group}, which comprised five individual datasets (i.e., CLEF'22, CLEF'23, MultiClaim, LESA, EnvClaims), and one combined \textit{general claim dataset}, and 2) a \textit{test group}, which consisted of two unseen datasets (i.e., NewsClaims and ClaimBuster)

In our experiments, we worked with seven diverse datasets, dividing each one into train, validation, and test sets. For three of these datasets, CLEF'22, CLEF'23, and LESA, predefined split ratios were utilized. The detailed number of samples for each dataset's train, validation and test set for each language is described in the Table \ref{tab:dataset-split-ratio} in the Appendix. For the remaining datasets, namely EnvClaims, NewsClaims, ClaimBuster, and MultiClaim, there were no default split ratios defined, therefore we determined the split ratios ourselves as 70\% of samples for training, 15\% for validation, and 15\% for testing.  

We performed two types of scenarios for experiments, where each experiment gradually increases the generality of the experiments. In our experiments, we trained models using the hyperparameters outlined in Table \ref{sec:appendix-implementation}.

\paragraph{In-distribution evaluation.}
In the first scenario, we conducted in-distribution evaluations to assess the models' performance when trained and tested within the same dataset. Specifically, we trained the models using the training set of a specific dataset, ensuring that they became familiar with the data distribution of that particular source. Subsequently, we evaluated the models' performance on the test set of the same dataset. We performed experiments on each dataset from the \textit{train group}. This scenario provides insights into how well the models generalize within the confines of a single dataset. 

\paragraph{Out-of-distribution evaluation.}
In the second scenario, we conducted an out-of-distribution evaluations to test the models' ability to generalize to unseen data sources. To achieve this, we trained the models on the \textit{train group}. Then, we subjected the models to testing using the \textit{test group} of previously unseen datasets. This scenario allowed us to gauge the models' adaptability and generalization capabilities beyond their original training domain. 

\paragraph{Evaluation metrics}
To determine the most effective models in our evaluation, we considered three key performance metrics: Balanced accuracy, recall, and F1-score. These metrics collectively provide a comprehensive assessment of a model's performance, capturing both its ability to make correct predictions and its effectiveness in identifying relevant information. By analyzing these metrics, we gained valuable insights into the overall effectiveness and robustness of the models under scrutiny. The balanced accuracy is defined as the average of recall obtained on each class ~\cite{Kelleher2015-hl}. To assess the performance of each model, we conduct a comparison of the F1-scores for both check-worthy and not check-worthy classes. All experiments were performed in a single run using a unique parameter set.


\section{Experiments and results}
In this section, we provide a summary of the outcomes from the conducted experiments, which can be categorized into two distinct parts: the results obtained from experiments with fine-tuned language models and those derived from experiments involving LLMs. 

\subsection{Fine-tuned language models}

\paragraph{In-distribution evaluation.}
Two sets of experiments were conducted: the first on each individual dataset, restricted solely to the English language, and the second conducted once again on each individual dataset but with an unrestricted multilingual approach spanning 10 languages.

\begin{figure*}
    \centering
    \includegraphics[width=1.0\textwidth] {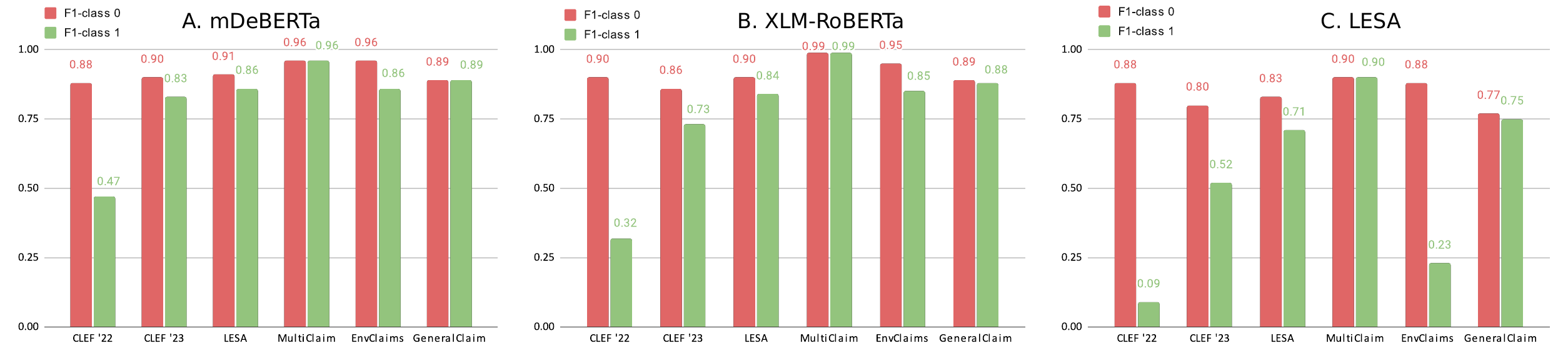}
    \caption{In-distribution evaluation of the \textbf{mDeBERTa} (A),  textbf{XLM-RoBERTa} (B) and \textbf{LESA} (C), conducted in a \textbf{multilingual} configuration in which datasets were unrestricted and comprised of all the 10 selected languages.}
    \label{fig:in-all}
\end{figure*}

\begin{figure*}[th]
    \centering
    \includegraphics[width=1.0\textwidth] {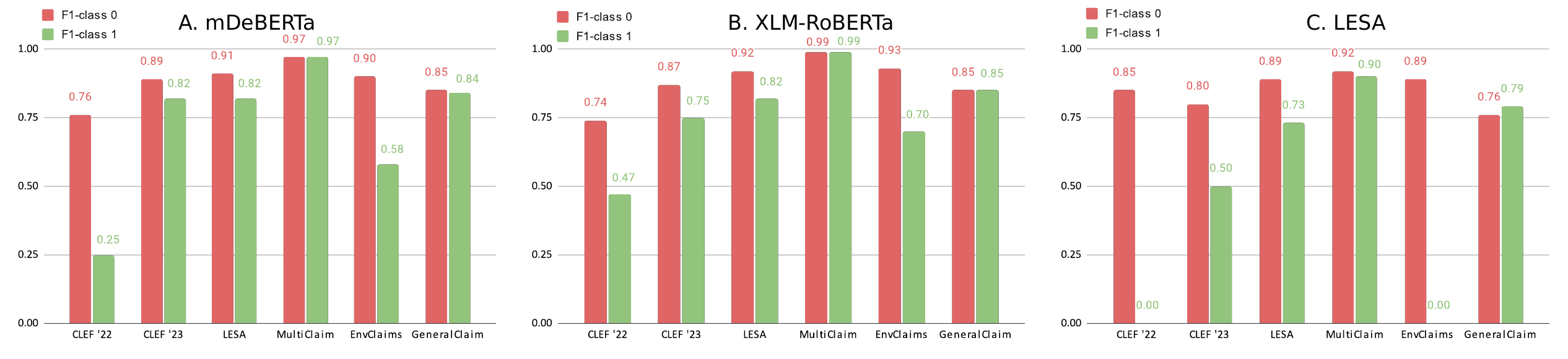}
    \caption{In-distribution evaluation of the \textbf{mDeBERTa} (A) models, \textbf{XLM-RoBERTa} (B) and \textbf{LESA} (C), conducted on a subset of the \textbf{English language} extracted from each tested dataset}
    \label{fig:in-en}
\end{figure*}

\textbf{(1) mDeBERTa excels in multilingual settings and LESA performs better in English-only environments}. In our experiments on individual datasets, we observed significant performance differences among the three models across both multilingual and restricted (English-only) datasets (see Figure~\ref{fig:in-all} and Figure~\ref{fig:in-en}). \textbf{(2)} \textit{Both mDeBERTa (A) and XLM-RoBERTa (B) consistently demonstrated strong performance across all datasets}, although all three models encountered challenges during training on the CLEF'22 dataset. In contrast, \textbf{(3)} \textit{LESA (C) model faced difficulties also with the EnvClaims dataset}, resulting in slightly lower performance compared to the other two models. This performance gap can be attributed to the fact that LESA model relies on the mBERT model, which is inherently inferior to the aforementioned models.

It is also important to emphasize that the models exhibited generally robust performance on individual datasets, a task considered comparatively less challenging than out-of-distribution evaluation. More details on experimental results are presented in Appendix~\ref{sec:appendix-in-dist}.


\paragraph{Out-of-distribution evaluation.}
In a more general scenario, models were trained on individual datasets and evaluated on two unseen datasets (NewsClaims and ClaimsBuster), with one of the evaluation datasets being our General Claim dataset. 

\begin{figure}
    \centering
    \includegraphics[width=0.45\textwidth] {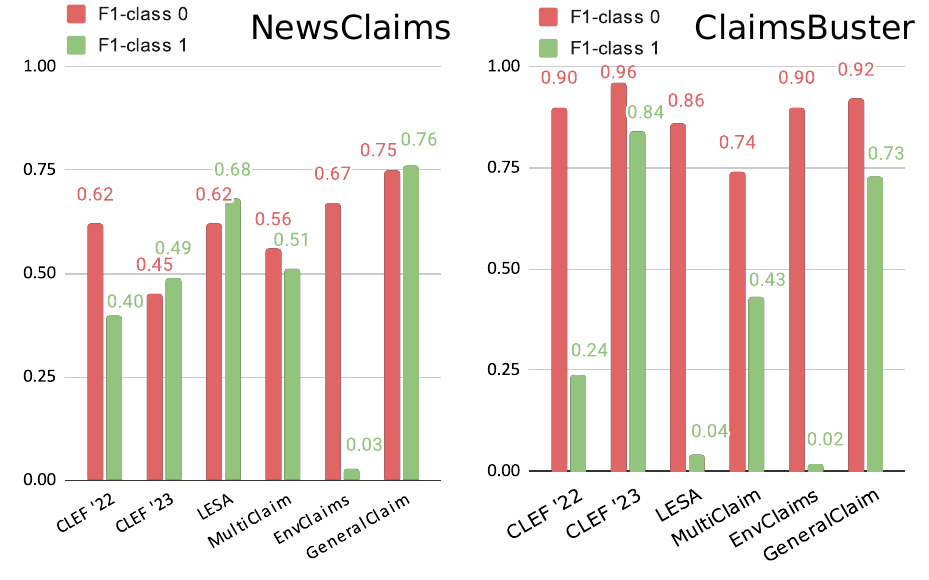}
    \caption{F1-score for \textbf{mDeBERTa} model, conducted in a \textbf{multilingual} configuration, trained on 6 selected datasets and tested in out-of-distribution scenario on NewsClaims and ClaimBuster datasets.}
    \label{fig:out-mdb-all}
\end{figure}

\begin{figure}
    \centering
    \includegraphics[width=0.45\textwidth] {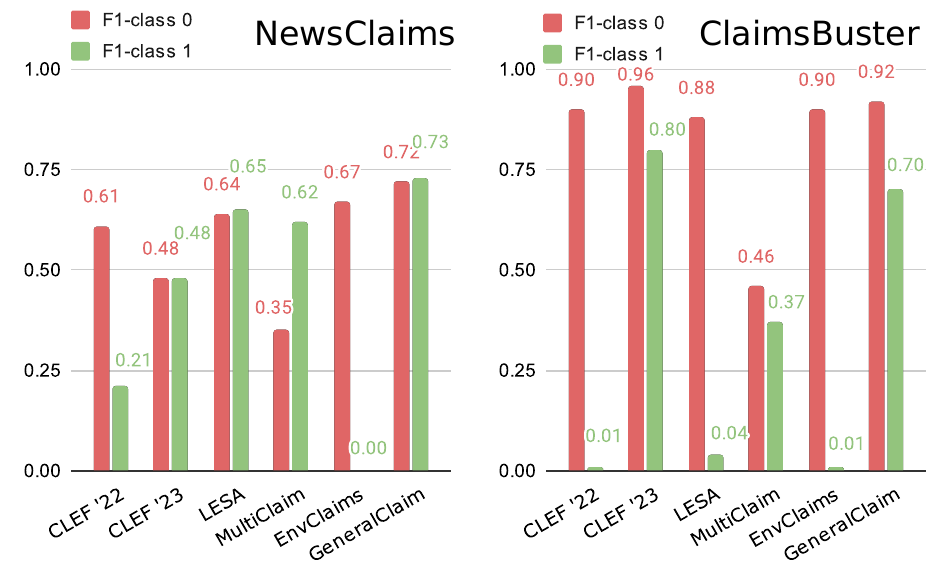}
    \caption{F1-score for \textbf{XLM-RoBERTa} model, conducted in a \textbf{multilingual} configuration, trained on 6 selected datasets and tested in out-of-distribution scenario on NewsClaims and ClaimBuster datasets.}
    \label{fig:out-xlm-all}
\end{figure}

\begin{figure}
    \centering
    \includegraphics[width=0.5\textwidth] {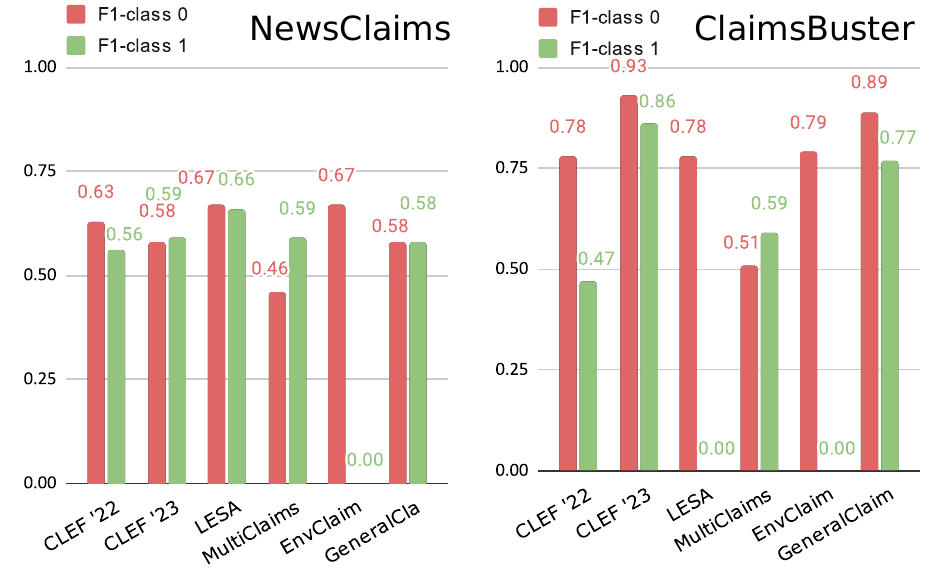}
    \caption{F1-score for \textbf{XLM-RoBERTa} model, trained on on a subset of the \textbf{English language} and tested in out-of-distribution scenario on NewsClaims and ClaimBuster datasets.}
    \label{fig:out-xlm-en}
\end{figure}

In terms of balanced accuracy, \textbf{(1) mDeBERTa was the best model on unrestricted multilingual datasets (91\%), which suggest that this model emerged as winner in our benchmark}. \textbf{(2)} \textit{XLM-RoBERTa model performed best on restricted dataset (90\%)} (see Figure~\ref{fig:out-xlm-en}). On this scenario, all three models struggled to adapt to EnvClaims, CLEF '22 and in some cases also LESA dataset. Also from the perspective of F1-score these two models showed the most balanced score of both classes (see Figure~\ref{fig:out-mdb-all} and~\ref{fig:out-xlm-all}. The \textbf{(3)} \textit{LESA model  performed least effective} (74\%), although it outperformed other two models on LESA dataset (58\%). On General claim dataset, containing multitude of topics, \textbf{(4)} \textit{both mDeBERTa (91\%) and XLM-RoBERTa (86\%) performed superior over the LESA model (74\%)}. For more details on experiments also on a datasets restricted to English-only, see Appendix~\ref{sec:appendix-out-dist}.

\subsection{Large Language Models}
Our deliberate choice involved selecting OpenAI models and we compare their performance against that of the Alpaca-LoRA and Mistral-7B models, which are presumed to be considerably smaller (though the exact sizes of OpenAI models remain undisclosed). To maintain consistency with our earlier experiments, we subjected these models to evaluation using the same set of evaluation datasets, however we refrained from fine-tuning these models in any manner. 

\begin{figure}
    \centering
    \includegraphics[width=0.45\textwidth] {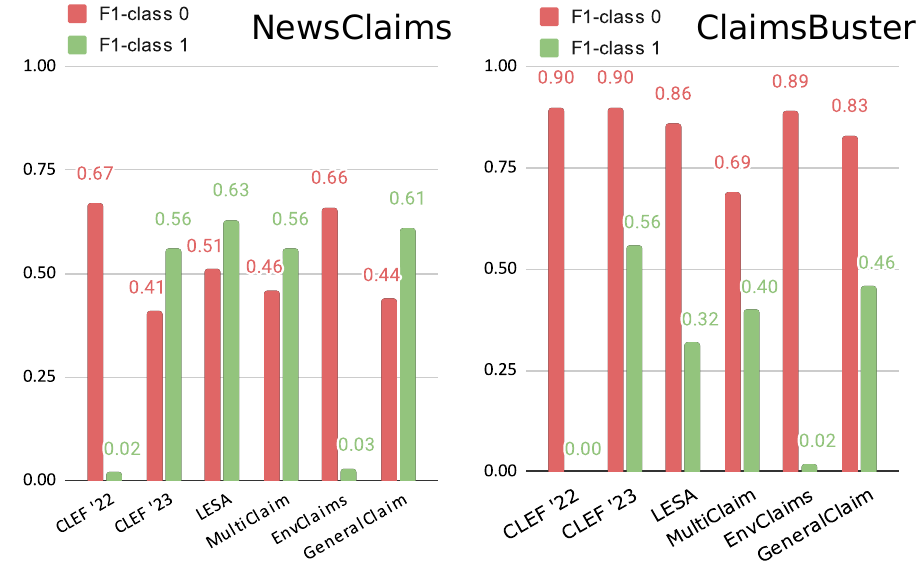}
    \caption{F1-score for \textbf{LESA} model, conducted in a \textbf{multilingual} configuration, trained on 6 selected datasets and tested in out-of-distribution scenario on NewsClaims and ClaimBuster datasets.}
    \label{fig:out-lesa-all}
\end{figure}

\begin{figure}
    \centering
    \includegraphics[width=0.5\textwidth] {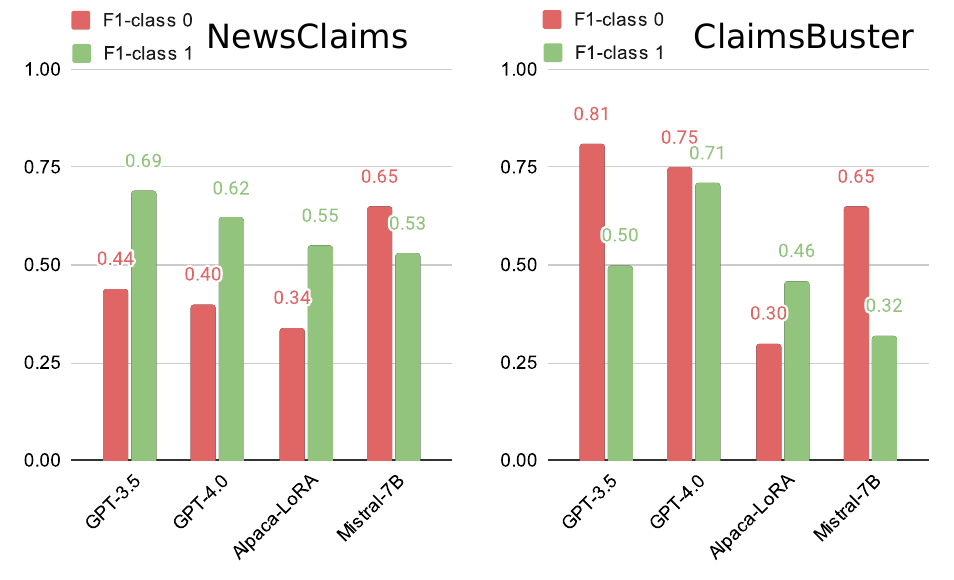}
    \caption{Out-of-distribution evaluation of the \textbf{LLM} models, tested on a subset of the \textbf{English} language extracted from each tested dataset.}
    \label{fig:out-LLM-en}
\end{figure}

\textbf{(1) GPT-4.0 exhibited superior performance on the multilingual dataset}, achieving a balanced accuracy of 0.78 compared to GPT-3.5's 0.58. Much like our evaluation of fine-tuned language models, our experiments here comprised of English only version and multilingual version of the dataset, results of which are presented in Tables \ref{tab:LLM-en} and \ref{tab:LLM-all} respectively. The results show that \textbf{(2)} \textit{on the NewsClaims dataset, GPT-4.0 generally outperformed GPT-3.5}, with higher balanced accuracy, recall, and F1-scores. Notably, \textbf{(3)} \textit{Alpaca-LoRA performed relatively poorly across both versions of the dataset}, with particularly low recall for negative class on the multilingual dataset with balanced accuracy of 0.07. \textbf{(4)} \textit{Mistral-7B showed competitive performance, especially in the NewsClaims dataset}, where it outperformed other models in balanced accuracy (0.60).

\textbf{(5) On the ClaimsBuster dataset, GPT-4.0 consistently outperformed GPT-3.5}, demonstrating higher balanced accuracy, recall, and F1-scores. Notably, GPT-4.0 achieved an impressive recall of 0.96 for positive class on the multilingual dataset. \textbf{(6)} \textit{Alpaca-LoRA again displayed suboptimal performance}, with a low recall for negative class in both english only and restricted datasets. \textbf{(7)} \textit{Mistral-7B exhibited mixed results}, with competitive balanced accuracy in the NewsClaims dataset but comparatively lower performance in the ClaimsBuster dataset.

\begin{figure}
    \centering
    \includegraphics[width=0.5\textwidth] {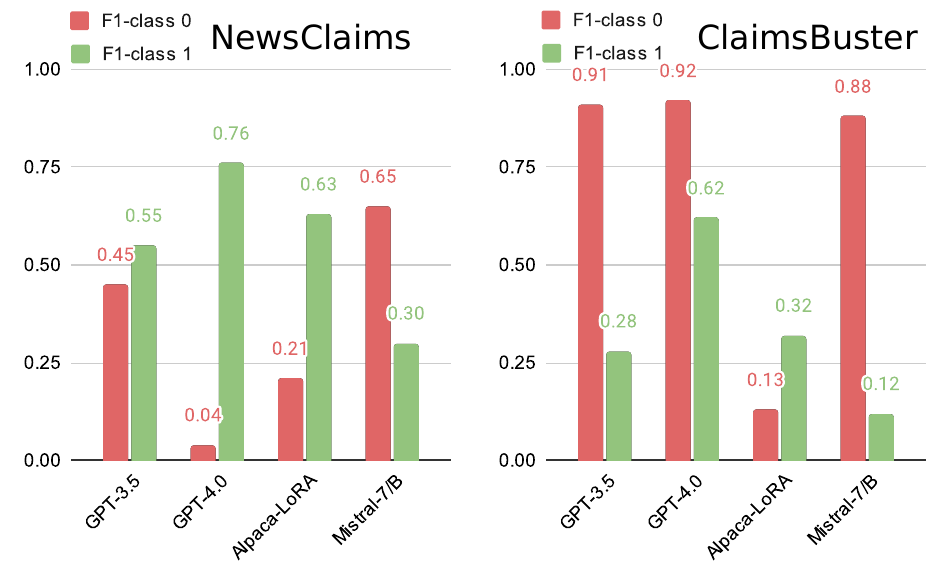}
    \caption{Out-of-distribution evaluation of the \textbf{LLM} models, conducted in a \textbf{multilingual} configuration in which datasets were unrestricted and comprised of all the 10 selected languages.}
    \label{fig:out-LLM-all}
\end{figure}

\textbf{(8) In contrast to fine-tuned language models, LLMs currently fall short in terms of precision}. It's worth noting that in this study, we did not delve into the potential of in-context learning for these models using a variety of prompts, even though the models possess this capability. We intend to explore this aspect in our future work.

\section{Discussion}
Comparing the results of in-distribution and out-of-distribution experiments scenarios for the F1-score of positive class, it can be observed, that \textit{mDeBERTa  achieves in most of the cases the best results among all the tested fine-tuned language models}. This is consistent with the results from prior works that report that mDeBERTa outperforms XLM-RoBERTa and mBERT architectures in downstream multilingual NLP task~\cite{DBLP:conf/iclr/HeGC23}. 

Comparing LLMs with fine-tuned language models in relation to the results obtained for multilingual data, the best results for both types of models are similar. The \textit{fine-tuned language models are performing slightly better}, which allows to conclude that the transfer of knowledge and fine tuning methods with appropriately selected training datasets provide greater opportunities to obtain better models for detection check-worthy claims than the use of general, not fine-tuned LLMs.

The comparison of LLMs reveals also, that their performance is in general better for English set than for multilingual test set. It is in accordance with observation for multilingual LLM, like mDeBERTaV3 or XLM-RoBERTa, where better results were reported for English than for other languages ~\cite{DBLP:conf/iclr/HeGC23} ~\cite{DBLP:conf/acl/ConneauKGCWGGOZ20}.

\section{Conclusion}
This work presents a comprehensive analysis of models for detecting check-worthy claims. We compared two types of models: fine-tuned language models and Large Language Models (LLMs). We demonstrate that it was challenging to conclusively determine whether the model size significantly influences performance. This conclusion was backed by the fact, that for the two types of models analyzed in this paper, the best results were very similar. We demonstrate that, through fine-tuning, better results (though only slightly better) can be obtained than from the best LLM, i.e., GPT4.0. However, fine-tuning methods necessitated meticulous data preparation, proving time-consuming and posing an additional challenge compared to equally accurate LLMs.

Future work will refer to in-context learning models and will focus on analyzing the impact of the prompt engineering. It will also involve fine-tuning of the in-context learning models on datasets and instructions related to check-worthy claims.


\clearpage

\section*{Limitations}
\label{sec:limitations}
The datasets used for fine-tuning are somewhat limited. The data used are multi-topical and multi-lingual, but its multi-thematic nature referred to 6 different topics (Environment, Health, Politics, Science, Sports, Entertainment) as in Table \ref{tab:datasets-topics}, and multilinguality to 10 different languages. Doubtless, the advantage of LLMs is their ability to make inferences for much more natural languages than 10, and in a much wider range of topics than the fine-tuned language models presented in this work.

The LLM models as mentioned were trained and tested in 10 languages, which is relatively high number compared to other models of this type ~\cite{DBLP:conf/clef/NakovBMAMCKZLSM22}. The different behavior of the models was observed depending on the specific language, hence an impact of language on the results was observed. Therefore, it should be expected that for languages that were not directly tested this impact will also occur. Therefore, it is also expected that the models will have lower performance for languages other than the 10 languages tested. The above statement applies in particular to LLM models (i.e. mDeBERTa, XLM-RoBERTa, LESA), although differences depending on the specific language were also observed for LLM models.

A similar limitation applies to the subject matter - the research was conducted on multi-topic texts relating to 6 thematic issues. It is expected that LLM models processing input texts beyond the mentioned thematic scope will obtain less performance of the results and this is clearly their limitation.

The models were tested on two datasets (NewsClaims and ClaimBuster), which were not used in any way for training. As the split ratio for testing set was 15\% and it was determined by the authors of this paper, it means that the validity and correctness of the results were limited to this type of testing. Another limitation was the selection of the prompt content for LLM models. They were examined in detail for one primary prompt with the following content \textit{``Does the input contain a check-worthy claim? Answer strictly in binary \textit{Yes} or \textit{No}''}.



\section*{Ethics Statement}
\label{sec:ethics}
In the pursuit of developing robust check-worthy claim detection models, ethical considerations are paramount to ensure responsible and accountable deployment. There is a potential concern that the AI models, if not transparently communicated, may generate uncertainty about their capabilities, leading to over-reliance on their use. Establishing mechanisms for auditability, technical robustness, and safety poses a significant challenge but is crucial to prevent unintended consequences. Moreover, there is a risk of negative discrimination against certain groups if biases are inadvertently embedded in the AI models. Misuse of these models is a pressing ethical issue, and safeguards must be implemented to prevent malicious activities. Additionally, the possibility of adversarial effects during technical faults, low accuracy, defects, or outages necessitates careful consideration to minimize potential harm to individuals. Striking a balance between technological advancement and ethical responsibility is imperative to foster trust and ensure the positive societal impact of check-worthy claim detection models.

In this work, we do not republish the datasets that we used in our study as it might mean the breach of the access restrictions imposed on some of the datasets. However, in our github repository we will publish the tools and mechanisms to process the original datasets into the format we used in our work, allowing the replication of our results. We have also performed a thorough ethical assessment of all its aspects (data, processes), using extended ethics checklist comprising 33 questions. This assessment yielded no concerns, nor pointed to any issues that needed to be resolved. This assessment procedure is proprietary to our organization and was created in a multi-stage process involving researchers, engineers, ethics experts, lawyers and other stakeholders.


\bibliography{anthology,custom}
\bibliographystyle{acl_natbib}

\appendix


\section{Selected publicly available datasets}
\label{sec:appendix-datasets}
For our study, we selected seven publicly available datasets. For three of these datasets, CLEF'22, CLEF'23, and LESA, predefined split ratios were utilized, with CLEF'22 having 42\% of samples for training, 48\% for validation, and 9\% for testing, CLEF'23 using 70\% of samples for training, 28\% for validation, and 2\% for testing, and LESA employing 81\% samples for training, 9\% for validation, and 10\% for testing. These split ratios varied across all the 10 languages contained in each dataset, therefore, for the sake of space, we are providing the mean split ratios, omitting the detailed calculations for each language and set. The detailed number of samples for each dataset's train, validation and test set for each language is described in Table \ref{tab:dataset-split-ratio}. For the remaining datasets, namely EnvClaims, NewsClaims, ClaimBuster, and MultiClaim, there were no default split ratios defined, therefore we determined the split ratios ourselves as 70\% of samples for training, 15\% for validation, and 15\% for testing.
Some of the datasets contained only a subset of the 10 languages we selected for our study. Therefore, we applied the translation of the English samples into each of the selected language using googletrans Python package version 3.1.0a0.

\begin{table}
    \centering
    \small
    \setlength\tabcolsep{4pt}
    \begin{tabular}{lccccccc}
        \hline
        \textbf{Dataset} & \textbf{Language} & \textbf{Train} & \textbf{Validation} & \textbf{Test} \\
        \hline
        \multirow{7}{*}{CLEF'22} 
        & English & 2122 & 2491 & 149 \\
        & Arabic & 2513 & 2999 & 682 \\
        & Bulgarian & 1871 & 2253 & 130 \\
        & Dutch & 923 & 1080 & 666 \\
        & Spanish & 4990 & 12500 & 5000 \\
        & Turkish & 2417 & 2862 & 303 \\
        & Slovak & 2122 & 1917 & 149 \\
        & Czech & 2122 & 1917 & 149 \\
        & Polish & 2122 & 1917 & 149 \\
        & Hungarian & 2122 & 1917 & 149 \\
        \hline
        \multirow{7}{*}{CLEF'23} 
        & English & 16876 & 6657 & 318 \\
        & Arabic & 16876 & 6657 & 318 \\
        & Bulgarian & 16876 & 6657 & 318 \\
        & Dutch & 16876 & 6657 & 318 \\
        & Spanish & 16876 & 6657 & 318 \\
        & Turkish & 16876 & 6657 & 318 \\
        & Slovak & 16876 & 6657 & 318 \\
        & Czech & 16876 & 6657 & 318 \\
        & Polish & 16876 & 6657 & 318 \\
        & Hungarian & 16876 & 6657 & 318 \\
        \hline
        \multirow{7}{*}{LESA} 
        & English & 31021 & 3629 & 3632 \\
        & Arabic & 31020 & 3629 & 3632 \\
        & Bulgarian & 31020 & 3629 & 3632 \\
        & Dutch & 31020 & 3629 & 3632 \\
        & Spanish & 31020 & 3629 & 3632 \\
        & Turkish & 31020 & 3629 & 3632 \\
        & Slovak & 31020 & 3629 & 3632 \\
        & Czech & 31020 & 3629 & 3632 \\
        & Polish & 31020 & 3629 & 3632 \\
        & Hungarian & 31020 & 3629 & 3632 \\
        \hline
    \end{tabular}
    \caption{Number of samples for train, validation, and test datasets for each language and each dataset.}
    \label{tab:dataset-split-ratio}
\end{table}

\textit{CLEF-CheckThat! Lab} is an annual evaluation initiative within the CLEF (Conference and Labs of the Evaluation Forum) framework, focusing on various tasks of identification of relevant claim in social media and web content. The lab challenges participants to develop algorithms and models capable of automatically assessing the check-worthiness of claims or statements present in online content, such as tweets, news articles, or social media posts. By determining the potential accuracy and credibility of these claims, CLEF-CheckThat! Lab contributes to the vital research field of misinformation detection, fact-checking, and credibility assessment in the context of the widespread dissemination of information through online platforms. Since 2021 the initiative annually publishes the datasets, each separated into multiple tasks. For our evaluation, we selected the CLEF2022-CheckThat! Lab dataset task 1~\cite{DBLP:conf/clef/NakovBMAMCKZLSM22}, which aims on prediction of  check-worthy claims in a Twitter stream posts, focusing on COVID-19 and politics in six languages: Arabic, Bulgarian, Dutch, English, Spanish, and
Turkish. We also selected the CLEF2023-CheckThat! Lab dataset task 1B~\cite{clef-checkthat:2023:task1}, focusing on Check-worthiness estimation from Multigenre (Unimodal) text - Tweets and/or transcriptions content in three languages: Arabic, English and Spanish. 

For purpose of our evaluation, we translated the English claims of the CLEF2023-CheckThat! Lab dataset into missing languages from CLEF2022-CheckThat! Lab dataset, namely Bulgarian, Dutch and Turkish. We also translated both datasets into the four other languages of V4 countries, namely Slovak, Czech, Polish and Hungarian. This way, we obtained two datasets comprising of ten languages: Arabic, Bulgarian, Dutch, English, Spanish, Turkish, Slovak, Czech, Polish and Hungarian.

\textit{LESA (2021)} (Linguistic Encapsulation and Semantic Amalgamation) is a generalised claim detection framework from online content which aims at assembling a source-independent generalized model that captures syntactic features through part-of-speech and dependency embeddings, as well as contextual features through a fine-tuned language model~\cite{gupta-etal-2021-lesa}. For the purpose of training the model, the authors assembled a generalized dataset for claim detection in online social media platforms containing 38,282 tweets from various sources~\cite{Coronavirus-Tweets, Coronavirus-(covid19)-Tweets, info:doi/10.2196/19273, DBLP:journals/corr/abs-2005-11177} and manually annotated it. The authors targeted and annotated only a subset of claims, i.e., factually verifiable claims and did not consider personal opinions, sarcastic comments, implicit claims, or claims existing in a sub-sentence or sub-clause level. Subsequently, they propose their own definition of claims and extrapolate the existing guidelines~\cite{DBLP:journals/corr/abs-2005-00033} to be more inclusive, nuanced and applicable to a diverse set of claims. Their definition for claims, adopted from Oxford dictionary, is to \textit{state or assert that something is the case, with or without providing evidence or proof}. Following the extrapolated guidelines, they re-annotated the tweets of~\cite{DBLP:journals/corr/abs-2005-00033} and CLEF-2020~\cite{DBLP:journals/corr/abs-2001-08546}.

For our evaluation, we employed both the LESA generalized dataset as well as the LESA language model for claim detection as a candidate model for the benchmark. We also incorporated the LESA dataset into our general claim dataset for general claim classification. To ensure the compatibility with the evaluation framework, we transled the dataset claims into nine additional languages to align the dataset with the languages featured in the CLEF-CheckThat! Lab datasets. The final dataset encompasing the following languages: Arabic, Bulgarian, Dutch, English, Spanish, Turkish, Slovak, Czech, Polish, and Hungarian.

\textit{NewsClaims (2022)} A Benchmark for Claim Detection from News with Background Knowledge~\cite{reddy2021newsclaims}. The benchmark was performed on the custom made dataset which specifically focuses on fact-checking claims made in news articles, enhancing the coverage of news-related content. Authors assembled a corpus with 889 claims over 143 diverse news articles, aimed to benchmark claim detection systems in emerging scenarios, comprising unseen topics with little or no training data. The dataset was also evaluated using the zero-shot and prompt-based models.

In our benchmark, we used the NewsClaims dataset as a part of evaluation, to have the reference to the results published in NewsClaims benchmark. To ensure a rigorous evaluation process, we translated the dataset claims into nine additional languages to align the dataset with the languages featured in the CLEF-CheckThat! Lab datasets. The final translated dataset encompassed Arabic, Bulgarian, Dutch, English, Spanish, Turkish, Slovak, Czech, Polish, and Hungarian.

\textit{MultiClaim dataset} ~\cite{pikuliak2023multilingual} is a part of the MultiClaim platform \cite{pikuliak2023multilingual}, designed to support the characterization and detection of various types of online antisocial behavior. It collects multimodal, multilingual context-rich data from multiple web sources, including news articles, blogs, and social media comments. The dataset is diverse and rich, supporting different content types and languages, as well as considering the credibility of authors. It is designed for easy extension with data-driven methods, enabling the handling of large amounts of unlabeled and dynamically evolving data. Furthermore, the dataset is continuously updated in real-time and supports various end-user services, making it a valuable resource for research on antisocial behavior detection and mitigation. It is an ongoing initiative which, in the time of writing this paper contained approx. 200,000 multi-topical fact-checking articles. Authors provide several baselines for these two tasks and evaluate them on the manually labeled part of the dataset. The dataset enables a number of additional tasks related to medical misinformation, such as misinformation characterisation studies or studies of misinformation diffusion between sources.

In order to align the MultiClaim (2022) dataset, composed solely of claims samples (positive class samples), with other datasets from our selection, we undertook a twofold approach. Initially, we restricted the dataset to 10K samples and filtered out languages not shared with other datasets, focusing exclusively on the ten languages employed in prior datasets. This strategic curation ensured compatibility across datasets in terms of size and used languages. Subsequently, we balanced the class distribution by integrating MultiClaim's positive class samples with synthetically generated negative class examples using Vicuna 7B LLM and analyzed them to ensure that they share the same characteristics as the original MultiClaim samples. To balance the number of samples across the languages we employed the translation of small number of missing samples from the English samples into a particular language.

\textit{Environmental claims dataset (2022)}~\cite{stammbach2023environmental} is dedicated to environmental claims and covers a wide range of environmental topics, enriching the general claim dataset with domain-specific information. The authors collected text from sustainability reports, earning calls, and annual reports of listed companies and annotated 2,647 examples by 16 in-domain experts. For constructing the dataset, the authors got the inspiration from the European Commission (EC), which defines such claims as follows: \textit{Environmental claims refer to the practice of suggesting or otherwise creating the impression (in the context of a commercial communication, marketing or advertising) that a product or a service is environmentally friendly (i.e., it has a positive impact on the environment) or is less damaging to the environment than competing goods or services.} \footnote{From the Commission Staff Working Document, Guidance on the implementation/application of Directive~\cite{Commission-Staff-Working-Document} on Unfair  Commercial practices, Brussels, 3 December 2009 SEC(2009) 1666. See section 2.5 on misleading environmental claims.}

\textit{ClaimBuster (2020)}
~\cite{arslan2020claimbuster} comprises 23,533 statements sourced from U.S. general election presidential debates, and each statement has been meticulously annotated by human coders. This dataset is invaluable for the development of computational methods aimed at discerning claims deserving of fact-checking within the vast array of digital and traditional media sources. Its extensive coverage and annotation make it a robust resource for researchers and developers working on the identification and evaluation of claims made during these debates, serving as a critical tool in the realm of information verification and accuracy in the realm of U.S. presidential elections.

\section{Fine-tuned language models}
\label{sec:appendix-finetuned}
\textit{The XLM-RoBERTa} is a multi-language model pretrained on 100 languages that performs well for many downstream tasks, including particularly well for low resource languages. The XLM-RoBERTa model was used in this work to create a classifier for detecting check-worthy claims in different languages (Arabic, Bulgarian, Dutch, English, Spanish, Turkish, Slovak, Czech, Polish and Hungarian). The model was trained on multilingual sets that included samples with check-worthy claims and samples without check-worthy claims. This method of training was adopted, to include the conclusions and outcomes of the paper~\cite{DBLP:conf/acl/ConneauKGCWGGOZ20}, where it was reported that in the classification task (detection of check-worthy claims is a typical classification task) the best results were obtained when training was performed from multiple languages. During the experiments, the XLM-RoBERTa-base architecture with 125 million parameters and the Flair library~\cite{flair_N19-4010}, version 0.7 were used. For the training and the inference of the model, we used the Flair Python framework of version 12.2.0 ~\cite{akbik2019flair}.

\textit{The LESA architecture} was also used in the experiments. The architecture consists of three basic elements: the POS (part-of-speech) module containing the BiLSTM neural network; DEP (dependency tree) module containing a transformer encoder; and transformer based BERT module~\cite{gupta-etal-2021-lesa}. The POS and DEP modules are used to extract syntactic features, and the BERT module is used to extract semantic features of the input text. The LESA architecture is characterized by the ability to detect claims in any types of texts, regardless of their source and regardless of their topic~\cite{gupta-etal-2021-lesa}. In the presented work, the LESA architecture was trained (fine-tuned) on various English-language collections in order to validate this solution in the tasks of detecting check-worthy claims in texts with a non-uniform subject and structure. To ensure compatibility in a multilingual setting, we made adaptations to our fine-tuned language models within the LESA architecture. Specifically, we replaced the BERT model in the original architecture with the multilingual mBERT.  For loading, training and inference of the model, we used the Tensorflow Python framework of version 2.10.0 ~\cite{tensorflow2015-whitepaper}, Keras framework of version 2.10.0 ~\cite{chollet2015keras} and spaCy NLP library of version 3.5.0 ~\cite{spacy2}.

\textit{The mDeBERTa} is a multilingual version of DeBERTa~\cite{DBLP:conf/iclr/HeLGC21}, which improves the BERT and RoBERTa models using disentangled attention and enhanced mask decoder. mDeBERTa~\cite{DBLP:conf/iclr/HeGC23} uses the same structure as DeBERTa and was trained with CC100 multilingual data. The mDeBERTa V3 base model comes with 12 layers and a hidden size of 768. It has 86M backbone parameters with a vocabulary containing 250K tokens which introduces 190M parameters in the Embedding layer. This model was trained using the 2.5T CC100 data as XLM-RoBERTa. With improvements of DeBERTa and with multilingual capabilities, mDeBERTa model out performs XLM-RoBERTa on XNLI benchmark~\cite{DBLP:conf/emnlp/ConneauRLWBSS18} with zero-shot cross-lingual transfer setting across 15 languages~\cite{DBLP:conf/iclr/HeGC23}.  For the training and the inference of the model, we used the Flair Python framework of version 12.2.0 ~\cite{akbik2019flair}.

\section{Large Language Models}
\label{sec:appendix-LLMs}
The results of the experiments on selected LLMs using selected datasets restricted to English language are contained in Table~\ref{tab:LLM-en}.

\begin{table}
    \centering
    \small
    \setlength\tabcolsep{3pt}
    \begin{tabular}{@{}cccccccc}
        \hline
        \textbf{Train Dataset} & \multicolumn{2}{c}{\textbf{Accuracy}} & \multicolumn{2}{c}{\textbf{Recall}} & \multicolumn{2}{c}{\textbf{F1-score}} \\
        \hline
        English &  &  & 0: & 1: & 0: & 1: \\
        \hline
        \multicolumn{6}{c}{\textbf{NewsClaims}} \\
        \hline
        \verb|GPT-3.5| & 0.61 & & 0.57 & 0.64 & 0.44 & 0.69 \\
        \verb|GPT-4.0| & 0.55 & & 0.57 & 0.52 & 0.40 & 0.62 \\
        \verb|Alpaca-LoRA| & 0.46 & & 0.27 & 0.65 & 0.34 & 0.55 \\
        \verb|Mistral-7B| & 0.60 & & 0.74 & 0.46 & 0.65 & 0.53 \\
        \hline
        \multicolumn{6}{c}{\textbf{CLEF'22}} \\
        \hline
        \verb|GPT-3.5| & 0.51 & & 0.51 & 0.51 & 0.61 & 0.35 \\
        \verb|GPT-4.0| & 0.67 & & 0.97 & 0.37 & 0.76 & 0.52 \\
        \verb|Alpaca-LoRA| & 0.51 & & 0.37 & 0.64 & 0.50 & 0.38 \\
        \verb|Mistral-7B| & 0.49 & & 0.52 & 0.46 & 0.61 & 0.33 \\
        \hline
        \multicolumn{6}{c}{\textbf{CLEF'23}} \\
        \hline
        \verb|GPT-3.5| & 0.65 & & 0.88 & 0.42 & 0.81 & 0.50 \\
        \verb|GPT-4.0| & 0.79 & & 0.61 & 0.96 & 0.75 & 0.71 \\
        \verb|Alpaca-LoRA| & 0.51 & & 0.16 & 0.85 & 0.25 & 0.49 \\
        \verb|Mistral-7B| & 0.53 & & 0.81 & 0.25 & 0.74 & 0.31\\
        \hline
        \multicolumn{6}{c}{\textbf{GeneralClaim}} \\
        \hline
        \verb|GPT-3.5| & 0.64 & & 0.79 & 0.48 & 0.66 & 0.58 \\
        \verb|GPT-4.0| & 0.74 & & 0.84 & 0.64 & 0.75 & 0.72 \\
        \verb|Alpaca-LoRA| & 0.46 & & 0.29 & 0.62 & 0.33 & 0.55 \\
        \verb|Mistral-7B| & 0.46 & & 0.67 & 0.25 & 0.53 & 0.33\\
        \hline
        \multicolumn{6}{c}{\textbf{ClaimsBuster}} \\
        \hline
        \verb|GPT-3.5| & 0.65 & & 0.88 & 0.42 & 0.81 & 0.50 \\
        \verb|GPT-4.0| & 0.79 & & 0.61 & 0.96 & 0.75 & 0.71 \\
        \verb|Alpaca-LoRA| & 0.48 & & 0.20 & 0.75 & 0.30 & 0.46 \\
        \verb|Mistral-7B| & 0.48 & & 0.65 & 0.31 & 0.65 & 0.32 \\
        \hline
    \end{tabular}
    \caption{Out-of-distribution evaluation of the \textbf{LLM} models, trained on a subset of the \textbf{English} language extracted from each tested dataset.}
    \label{tab:LLM-en}
\end{table}

The results of the experiments on selected LLMs using selected unrestricted datasets spanning 10 language are contained in Table~\ref{tab:LLM-all}.

\begin{table}
    \centering
    \small
    \setlength\tabcolsep{3pt}
    \begin{tabular}{@{}cccccccc}
        \hline
        \textbf{Train Dataset} & \multicolumn{2}{c}{\textbf{Accuracy}} & \multicolumn{2}{c}{\textbf{Recall}} & \multicolumn{2}{c}{\textbf{F1-score}} \\
        \hline
        Multilingual &  &  & 0: & 1: & 0: & 1: \\
        \hline
        \multicolumn{6}{c}{\textbf{NewsClaims}} \\
        \hline
        \verb|GPT-3.5| & 0.51 & & 0.53 & 0.49 & 0.45 & 0.55 \\
        \verb|GPT-4.0| & 0.51 & & 0.02 & 1.0 & 0.04 & 0.76 \\
        \verb|Alpaca-LoRA| & 0.50 & & 0.14 & 0.86 & 0.21 & 0.63 \\
        \verb|Mistral-7B| & 0.54 & & 0.87 & 0.20 & 0.65 & 0.30 \\
        \hline
        \multicolumn{6}{c}{\textbf{CLEF'22}} \\
        \hline
        \verb|GPT-3.5| & 0.61 & & 0.71 & 0.51 & 0.77 & 0.39 \\
        \verb|GPT-4.0| & 0.63 & & 0.35 & 0.90 & 0.51 & 0.40 \\
        \verb|Alpaca-LoRA| & 0.53 & & 0.15 & 0.90 & 0.25 & 0.45 \\
        \verb|Mistral-7B| & 0.54 & & 0.84 & 0.23 & 0.78 & 0.29\\
        \hline
        \multicolumn{6}{c}{\textbf{CLEF'23}} \\
        \hline
        \verb|GPT-3.5| & 0.57 & & 0.93 & 0.20 & 0.79 & 0.30 \\
        \verb|GPT-4.0| & 0.80 & & 0.74 & 0.85 & 0.81 & 0.72 \\
        \verb|Alpaca-LoRA| & 0.54 & & 0.17 & 0.90 & 0.28 & 0.51 \\
        \verb|Mistral-7B| & 0.52 & & 0.95 & 0.09 & 0.78 & 0.14\\
        \hline
        \multicolumn{6}{c}{\textbf{GeneralClaim}} \\
        \hline
        \verb|GPT-3.5| & 0.54 & & 0.84 & 0.24 & 0.65 & 0.34 \\
        \verb|GPT-4.0| & 0.69 & & 0.80 & 0.57 & 0.72 & 0.65 \\
        \verb|Alpaca-LoRA| & 0.47 & & 0.13 & 0.80 & 0.20 & 0.61 \\
        \verb|Mistral-7B| & 0.52 & & 0.88 & 0.15 & 0.64 & 0.24\\
        \hline
        \multicolumn{6}{c}{\textbf{ClaimsBuster}} \\
        \hline
        \verb|GPT-3.5| & 0.58 & & 0.96 & 0.19 & 0.91 & 0.28 \\
        \verb|GPT-4.0| & 0.78 & & 0.92 & 0.63 & 0.92 & 0.62 \\
        \verb|Alpaca-LoRA| & 0.52 & & 0.07 & 0.97 & 0.13 & 0.32 \\
        \verb|Mistral-7B| & 0.52 & & 0.95 & 0.08 & 0.88 & 0.12\\
        \hline
    \end{tabular}
    \caption{Out-of-distribution evaluation of the \textbf{LLM} models, trained on \textbf{multilingual} configuration comprising of all the 10 selected languages from all tested datasets.}
    \label{tab:LLM-all}
\end{table}

For the ChatGPT, the  GPT-3.5-turbo-0613 API was applied in this work ~\cite{DBLP:journals/corr/abs-2303-18223}, and for the GPT-4.0, which is the next generation of the GPT architecture and outperforms all previous GPT models ~\cite{DBLP:journals/corr/abs-2303-18223, DBLP:journals/corr/abs-2303-12712}. The API for GPT-4.0 released by OpenAI and applied in the work was gpt-4-0613.

\section{Implementation details}
\label{sec:appendix-implementation}

Experiments were performed in the Python programming language. All the datasets were processed using the Pandas Python package version 1.5.3 and stored in the csv data format. The smaller LLM models were fine-tuned using both Tensorflow Python framework of version 2.10.0 ~\cite{tensorflow2015-whitepaper} and  Flair Python framework of version 12.2.0 ~\cite{akbik2019flair}. Hyperparameters for the fine-tuning of smaller LLMs are described in Table \ref{tab:hyperparams}. For accessing OpenAI models we used OpenAI API ~\cite{openaiapi}. To perform experiments on Alpaca-LoRA ~\cite{Alpaca-LoRA} and Mistral-7B ~\cite{jiang2023mistral} we used huggingface models and run them on a local server. Performance metrics (balanced accuracy, recall, and F1-score) were computed using classification\_report function from the scikit-learn Python library of version 1.2.1 ~\cite{pedregosa2011scikit}. All the results were saved in the xlsl format again using the Pandas Python package.

\begin{table}
    \centering
    \small
    \setlength\tabcolsep{2pt}
    \begin{tabular}{lccc}
        \hline
        \textbf{Hyperparameter} & \textbf{mDeBERTa} & \textbf{XLM-R} & \textbf{LESA} \\
        \hline
        Batch size & 32 & 32 & 32 \\
        Learning rate & 3e-3 & 3e-3 & 3e-5\\
        Epochs & 5 & 5 & 5\\
        Optimizer & Adam & Adam & Adam \\
        Patience & 1 & 1 & - \\
        Pre-processing & True & True & True \\
        \hline
    \end{tabular}
    \caption{Hyperparamters used for fine-tuning the smaller LLMs in In-distribution as well as Out-of-distribution scenarios.}
    \label{tab:hyperparams}
\end{table}

\section{In-Distribution evaluation}
\label{sec:appendix-in-dist}
In the In-Distribution Evaluation scenario, we analyze how our model performs when presented with data that closely resembles the training data. In this experiments we also show, that we replicated the results of the models we use in our study. Specifically, we replicated the results of the LESA model trained on the LESA dataset~\cite{gupta-etal-2021-lesa}, where the authors reported Macro-F1 score of 0.79. In our evaluation, we obtained a result of 0.79, calculated as the balanced accuracy, shown in the Table~\ref{fig:in-en} (C). We additionally conducted experiments to enable a comparison with the existing baselines of CLEF2022-CheckThat! Lab dataset~\cite{DBLP:conf/clef/NakovBMAMCKZLSM22}\footnote{CLEF2022-CheckThat! Lab Leaderboards: \url{http://shorturl.at/dzX28}}. The referenced study conducted evaluations for each individual language, resulting in a range of scores from 0.2 to 0.65. In our experiments, using the same dataset with GeneralClaim languages, we achieved F1-scores (positive class) of 0.47 for the mDeBERTa model, and 0.32 and 0.09 for the XLM-R and LESA models, respectively. The results reflect the fact that  models were trained on general-case data, instead of narrow sub-problem. We achieved superior performance compared to the models evaluated in~\cite{stammbach2023environmental} on their Environmental Claims dataset. The original paper reported an accuracy of 0.917 (91.7) on the test set, while both our mDeBERTa and XLM-RoBERTa models attained a higher accuracy of 0.93.

\begin{table}
    \centering
    \small
    \setlength\tabcolsep{3pt}
    \begin{tabular}{@{}cccccccc}
        \hline
        \textbf{Train Dataset} & \multicolumn{2}{c}{\textbf{Accuracy}} & \multicolumn{2}{c}{\textbf{Recall}} & \multicolumn{2}{c}{\textbf{F1-score}} \\
        \hline
        English &  &  & 0: & 1: & 0: & 1: \\
        \hline
        \verb|CLEF'22| & 0.51 & & 0.78 & 0.23 & 0.76 & 0.25 \\
        \verb|CLEF'23| & 0.85 & & 0.96 & 0.74 & 0.89 & 0.82 \\
        \verb|LESA| & 0.86 & & 0.94 & 0.77 & 0.91 & 0.82 \\
        \verb|NewsClaims| & 0.55 & & 0.19 & 0.90 & 0.29 & 0.69 \\
        \verb|MultiClaim| & 0.97 & & 1.00 & 0.94 & 0.97 & 0.97 \\
        \verb|EnvClaims| & 0.84 & & 0.91 & 0.56 & 0.90 & 0.58 \\
        \verb|GeneralClaim| & 0.85 & & 0.93 & 0.77 & 0.85 & 0.84 \\
        \hline
    \end{tabular}
    \caption{In-distribution evaluation of the \textbf{mDeBERTa} model trained  on a subset of the \textbf{English} language extracted from each tested dataset.}
    \label{tab:in-mde-en}
\end{table}

\begin{table}
    \centering
    \small
    \setlength\tabcolsep{3pt}
    \begin{tabular}{@{}cccccccc}
        \hline
        \textbf{Train Dataset} & \multicolumn{2}{c}{\textbf{Accuracy}} & \multicolumn{2}{c}{\textbf{Recall}} & \multicolumn{2}{c}{\textbf{F1-score}} \\
        \hline
        Multilingual &  &  & 0: & 1: & 0: & 1: \\
        \hline
        \verb|CLEF'22| & 0.66 &  & 0.91 & 0.41 & 0.88 & 0.47 & \\
        \verb|CLEF'23| & 0.86 &  & 0.98 & 0.73 & 0.90 & 0.83 & \\
        \verb|LESA| & 0.89 &  & 0.93 & 0.84 & 0.91 & 0.86 & \\
        \verb|NewsClaims| & 0.76 &  & 0.69 & 0.83 & 0.74 & 0.79 & \\
        \verb|MultiClaim| & 0.96 &  & 1.00 & 0.92 & 0.96 & 0.96 & \\
        \verb|EnvClaims| & 0.94 &  & 0.94 & 0.93 & 0.96 & 0.86 & \\
        \verb|GeneralClaim| & 0.90 &  & 0.93 & 0.86 & 0.89 & 0.89 & \\
        \hline
    \end{tabular}
    \caption{In-distribution evaluation of the \textbf{mDeBERTa} model trained on \textbf{multilingual} configuration comprising of all the 10 selected languages from all tested datasets.}
    \label{tab::in-mde-all}
\end{table}

\begin{table}
    \centering
    \small
    \setlength\tabcolsep{3pt}
    \begin{tabular}{@{}cccccccc}
        \hline
        \textbf{Train Dataset} & \multicolumn{2}{c}{\textbf{Accuracy}} & \multicolumn{2}{c}{\textbf{Recall}} & \multicolumn{2}{c}{\textbf{F1-score}} \\
        \hline
        English &  &  & 0: & 1: & 0: & 1: \\
        \hline
        \verb|CLEF'22| & 0.63 &  & 0.67 & 0.59 & 0.74 & 0.47 \\
        \verb|CLEF'23| & 0.81 &  & 0.98 & 0.63 & 0.87 & 0.75 \\
        \verb|LESA| & 0.86 &  & 0.96 & 0.75 & 0.92 & 0.82 \\
        \verb|NewsClaims| & 0.42 &  & 0.26 & 0.58 & 0.30 & 0.52 \\
        \verb|MultiClaim| & 0.99 &  & 1.00 & 0.97 & 0.99 & 0.99 \\
        \verb|EnvClaims| & 0.82 &  & 0.92 & 0.71 & 0.93 & 0.70 \\
        \verb|GeneralClaim| & 0.86 &  & 0.92 & 0.79 & 0.85 & 0.85 \\
        \hline
    \end{tabular}
    \caption{In-distribution evaluation of the \textbf{XLM-RoBERTa} model trained on a subset of the \textbf{English} language extracted from each tested dataset.}
    \label{tab:in-xlm-en}
\end{table}

\begin{table}
    \centering
    \small
    \setlength\tabcolsep{3pt}
    \begin{tabular}{@{}cccccccc}
        \hline
        \textbf{Train Dataset} & \multicolumn{2}{c}{\textbf{Accuracy}} & \multicolumn{2}{c}{\textbf{Recall}} & \multicolumn{2}{c}{\textbf{F1-score}} \\
        \hline
        Multilingual &  &  & 0: & 1: & 0: & 1: \\
        \hline
        \verb|CLEF'22| & 0.59 &  & 0.99 & 0.19 & 0.90 & 0.32 \\
        \verb|CLEF'23| & 0.79 &  & 0.97 & 0.60 & 0.86 & 0.73 \\
        \verb|LESA| & 0.87 &  & 0.91 & 0.82 & 0.90 & 0.84 \\
        \verb|NewsClaims| & 0.76 &  & 0.62 & 0.90 & 0.71 & 0.80 \\
        \verb|MultiClaim| & 0.99 &  & 1.00 & 0.98 & 0.99 & 0.99 \\
        \verb|EnvClaims| & 0.92 &  & 0.93 & 0.90 & 0.95 & 0.85 \\
        \verb|GeneralClaim| & 0.89 &  & 0.93 & 0.84 & 0.89 & 0.88 \\
        \hline
    \end{tabular}
    \caption{In-distribution evaluation of the \textbf{XLM-RoBERTa} model trained on \textbf{multilingual} configuration comprising of all the 10 selected languages from all tested datasets.}
    \label{tab:in-xlm-all}
\end{table}

\begin{table}
    \centering
    \small
    \setlength\tabcolsep{3pt}
    \begin{tabular}{@{}cccccccc}
        \hline
        \textbf{Train Dataset} & \multicolumn{2}{c}{\textbf{Accuracy}} & \multicolumn{2}{c}{\textbf{Recall}} & \multicolumn{2}{c}{\textbf{F1-score}} \\
        \hline
        English &  &  & 0: & 1: & 0: & 1: \\
        \hline
        \verb|CLEF'22| & 0.50 &  & 1.00 & 0.00 & 0.85 & 0.00 \\
        \verb|CLEF'23| & 0.66 &  & 0.98 & 0.34 & 0.80 & 0.50 \\
        \verb|LESA| & 0.79 &  & 0.96 & 0.62 & 0.89 & 0.73 \\
        \verb|NewsClaims| & 0.61 &  & 0.56 & 0.66 & 0.57 & 0.65 \\
        \verb|MultiClaim| & 0.91 &  & 0.96 & 0.86 & 0.92 & 0.90 \\
        \verb|EnvClaims| & 0.50 &  & 1.00 & 0.00 & 0.89 & 0.00 \\
        \verb|GeneralClaim| & 0.77 &  & 0.76 & 0.78 & 0.76 & 0.79 \\
        \hline
    \end{tabular}
    \caption{In-distribution evaluation of the \textbf{LESA} model trained on a subset of the \textbf{English} language extracted from each tested dataset.}
    \label{tab:in-lesa-en}
\end{table}

\begin{table}
    \centering
    \small
    \setlength\tabcolsep{3pt}
    \begin{tabular}{@{}cccccccc}
        \hline
        \textbf{Train Dataset} & \multicolumn{2}{c}{\textbf{Accuracy}} & \multicolumn{2}{c}{\textbf{Recall}} & \multicolumn{2}{c}{\textbf{F1-score}} \\
        \hline
        Multilingual &  &  & 0: & 1: & 0: & 1: \\
        \hline
        \verb|CLEF'22| & 0.52 &  & 0.98 & 0.05 & 0.88 & 0.09 \\
        \verb|CLEF'23| & 0.67 &  & 0.96 & 0.37 & 0.80 & 0.52 \\
        \verb|LESA| & 0.77 &  & 0.86 & 0.67 & 0.83 & 0.71 \\
        \verb|NewsClaims| & 0.60 &  & 0.31 & 0.88 & 0.43 & 0.70 \\
        \verb|MultiClaim| & 0.90 &  & 0.90 & 0.90 & 0.90 & 0.90 \\
        \verb|EnvClaims| & 0.56 &  & 0.97 & 0.14 & 0.88 & 0.23 \\
        \verb|GeneralClaim| & 0.76 &  & 0.82 & 0.70 & 0.77 & 0.75 \\
        \hline
    \end{tabular}
    \caption{In-distribution evaluation of the \textbf{LESA} model trained on \textbf{multilingual} configuration comprising of all the 10 selected languages from all tested datasets.}
    \label{tab:in-lesa-all}
\end{table}

\section{Out-of-Distribution evaluation}
\label{sec:appendix-out-dist}
The Out-of-Distribution Evaluation scenario explores the model's capacity to generalize to unfamiliar data. We trained models separately on each individual dataset and assessed their performance on two unseen datasets (NewsClaims and ClaimsBuster). This approach allows us to gauge the models' ability to generalize across different topics. In cases where train and unseen datasets share the same topic, the similarity of the training data will of course play a certain role, however, after the examination of the results we can assert, that various models perform with drastically different accuracy in some cases. This difference can be considered as the measure of generalization. In the main paper body, we've addressed the overall results of our study. Here, we discuss the specific findings from three additional figures that focus on the outcomes derived from a restricted dataset limited to English-only content. 

With our General claim dataset, which is composed of multiple of topics, we are testing a special case, where the two datasets are guarantee to share at least one topic. This way we can measure more specifically the ability of the model to generalize within multiple topics. By testing it on specific topic (Politics in case of both unseen datasets), we are essentially evaluating, how well the model can cover one of the topics it was trained on and thus ascertain the quality of the model generalization capabilities.

\begin{figure}
    \centering
    \includegraphics[width=0.45\textwidth] {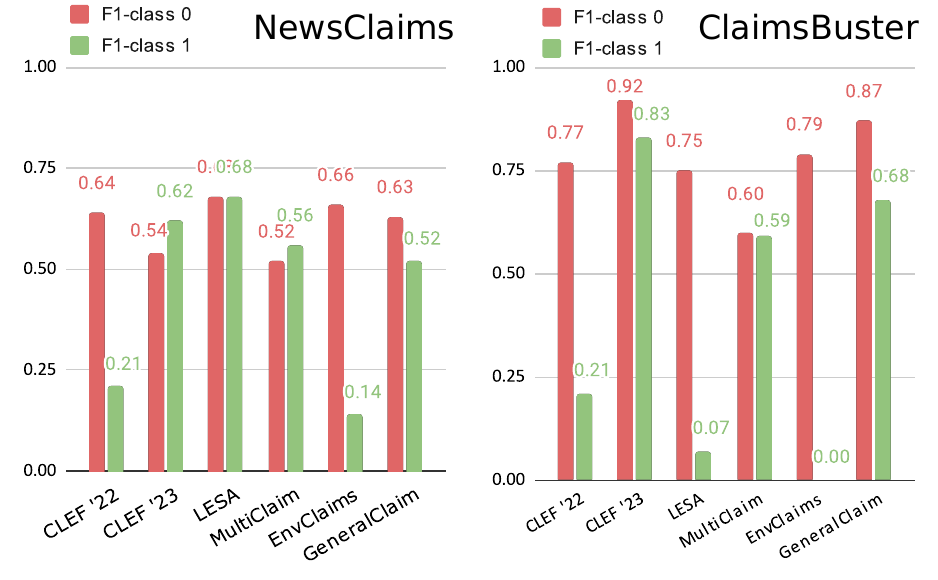}
    \caption{F1-score for \textbf{mDeBERTa} model, trained on on a subset of the \textbf{English language} and tested in out-of-distribution scenario on NewsClaims and ClaimBuster datasets.}
    \label{fig:out-mdb-en}
\end{figure}

\begin{figure}
    \centering
    \includegraphics[width=0.5\textwidth] {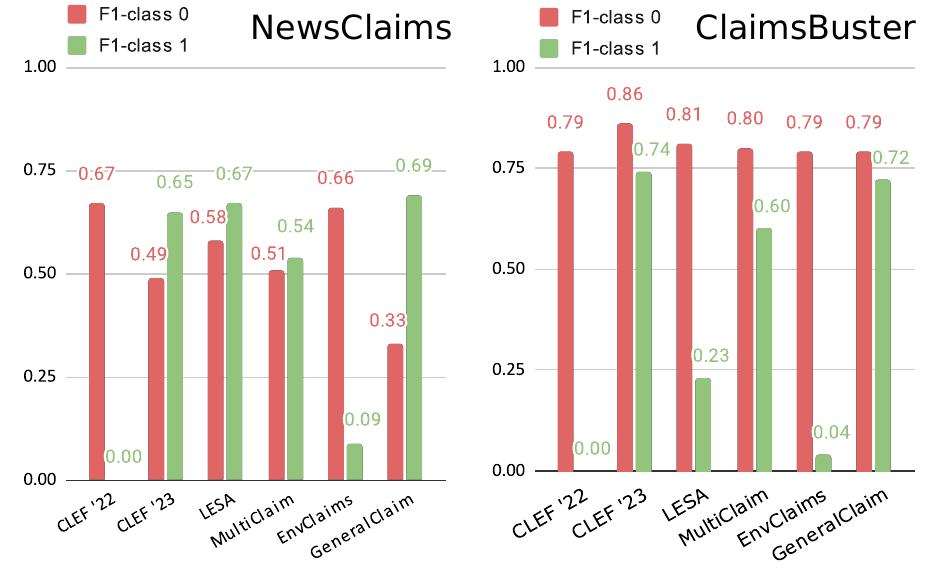}
    \caption{F1-score for \textbf{LESA} model, trained on on a subset of the \textbf{English language} and tested in out-of-distribution scenario on NewsClaims and ClaimBuster datasets.}
    \label{fig:out-lesa-en}
\end{figure}

\begin{table}
    \centering
    \small
    \setlength\tabcolsep{3pt}
    \begin{tabular}{@{}cccccccc}
        \hline
        \textbf{Train Dataset} & \multicolumn{2}{c}{\textbf{Accuracy}} & \multicolumn{2}{c}{\textbf{Recall}} & \multicolumn{2}{c}{\textbf{F1-score}} \\
        \hline
        English &  &  & 0: & 1: & 0: & 1: \\
        \hline
        \multicolumn{6}{c}{\textbf{NewsClaims}} \\
        \hline
        \verb|CLEF'22| & 0.51 & & 0.88 & 0.13 & 0.64 & 0.21 \\
        \verb|CLEF'23| & 0.59 & & 0.49 & 0.68 & 0.54 & 0.62 \\
        \verb|LESA| & 0.69 & & 0.69 & 0.68 & 0.68 & 0.68 \\
        \verb|NewsClaims| & 0.55 & & 0.19 & 0.90 & 0.29 & 0.69 \\
        \verb|MultiClaim| & 0.55 & & 0.54 & 0.55 & 0.52 & 0.56 \\
        \verb|EnvClaims| & 0.52 & & 0.95 & 0.08 & 0.66 & 0.14 \\
        \verb|GeneralClaim| & 0.60 & & 0.78 & 0.42 & 0.63 & 0.52 \\
        \hline
        \multicolumn{6}{c}{\textbf{CLEF'22}} \\
        \hline
        \verb|CLEF'22| & 0.51 & & 0.88 & 0.13 & 0.80 & 0.18 \\
        \verb|CLEF'23| & 0.63 & & 0.39 & 0.87 & 0.54 & 0.49 \\
        \verb|LESA| & 0.64 & & 0.32 & 0.95 & 0.48 & 0.49 \\
        \verb|NewsClaims| & 0.51 & & 0.07 & 0.95 & 0.13 & 0.42 \\
        \verb|MultiClaim| & 0.55 & & 0.28 & 0.82 & 0.42 & 0.43 \\
        \verb|EnvClaims| & 0.51 & & 0.92 & 0.10 & 0.82 & 0.15 \\
        \verb|GeneralClaim| & 0.55 & & 0.41 & 0.69 & 0.54 & 0.41 \\
        \hline
        \multicolumn{6}{c}{\textbf{CLEF'23}} \\
        \hline
        \verb|CLEF'22| & 0.56 & & 0.96 & 0.16 & 0.80 & 0.25 \\
        \verb|CLEF'23| & 0.85 & & 0.96 & 0.74 & 0.89 & 0.82 \\
        \verb|LESA| & 0.50 & & 0.98 & 0.02 & 0.79 & 0.04 \\
        \verb|NewsClaims| & 0.54 & & 0.17 & 0.90 & 0.27 & 0.59 \\
        \verb|MultiClaim| & 0.77 & & 0.67 & 0.76 & 0.74 & 0.63 \\
        \verb|EnvClaims| & 0.50 & & 1.00 & 0.00 & 0.80 & 0.00 \\
        \verb|GeneralClaim| & 0.80 & & 0.94 & 0.66 & 0.86 & 0.75 \\
        \hline
        \multicolumn{6}{c}{\textbf{GeneralClaim}} \\
        \hline
        \verb|CLEF'22| & 0.55 & & 0.90 & 0.20 & 0.64 & 0.31 \\
        \verb|CLEF'23| & 0.66 & & 0.86 & 0.46 & 0.71 & 0.58 \\
        \verb|LESA| & 0.64 & & 0.91 & 0.36 & 0.71 & 0.50 \\
        \verb|NewsClaims| & 0.54 & & 0.15 & 0.93 & 0.24 & 0.70 \\
        \verb|MultiClaim| & 0.76 & & 0.75 & 0.77 & 0.75 & 0.76 \\
        \verb|EnvClaims| & 0.49 & & 0.96 & 0.02 & 0.65 & 0.03 \\
        \verb|GeneralClaim| & 0.85 & & 0.93 & 0.77 & 0.85 & 0.84 \\
        \hline
        \multicolumn{6}{c}{\textbf{ClaimsBuster}} \\
        \hline
        \verb|CLEF'22| & 0.53 & & 0.93 & 0.13 & 0.77 & 0.21 \\
        \verb|CLEF'23| & 0.87 & & 0.95 & 0.78 & 0.92 & 0.83 \\
        \verb|LESA| & 0.48 & & 0.92 & 0.04 & 0.75 & 0.07 \\
        \verb|NewsClaims| & 0.47 & & 0.07 & 0.87 & 0.13 & 0.49 \\
        \verb|MultiClaim| & 0.65 & & 0.47 & 0.83 & 0.60 & 0.59 \\
        \verb|EnvClaims| & 0.50 & & 1.00 & 0.00 & 0.79 & 0.00 \\
        \verb|GeneralClaim| & 0.76 & & 0.95 & 0.57 & 0.87 & 0.68 \\
        \hline
    \end{tabular}
    \caption{Out-of-distribution evaluation of the \textbf{mDeBERTa} model trained on a subset of the English language.}
    \label{tab:out-mdb-en}
\end{table}

\begin{table}
    \centering
    \small
    \setlength\tabcolsep{3pt}
    \begin{tabular}{@{}cccccccc}
        \hline
        \textbf{Train Dataset} & \multicolumn{2}{c}{\textbf{Accuracy}} & \multicolumn{2}{c}{\textbf{Recall}} & \multicolumn{2}{c}{\textbf{F1-score}} \\
        \hline
        Multilingual &  &  & 0: & 1: & 0: & 1: \\
        \hline
        \multicolumn{6}{c}{\textbf{NewsClaims}} \\
        \hline
        \verb|CLEF'22| & 0.54 & & 0.77 & 0.31 & 0.62 & 0.40 \\
        \verb|CLEF'23| & 0.47 & & 0.43 & 0.50 & 0.45 & 0.49 \\
        \verb|LESA| & 0.66 & & 0.57 & 0.74 & 0.62 & 0.68 \\
        \verb|NewsClaims| & 0.79 & & 0.74 & 0.83 & 0.77 & 0.79 \\
        \verb|MultiClaim| & 0.54 & & 0.59 & 0.49 & 0.56 & 0.51 \\
        \verb|EnvClaims| & 0.51 & & 1.00 & 0.02 & 0.67 & 0.03 \\
        \verb|GeneralClaim| & 0.76 & & 0.79 & 0.73 & 0.75 & 0.76 \\
        \hline
        \multicolumn{6}{c}{\textbf{CLEF'22}} \\
        \hline
        \verb|CLEF'22| & 0.66 & & 0.91 & 0.41 & 0.88 & 0.47 \\
        \verb|CLEF'23| & 0.72 & & 0.70 & 0.74 & 0.79 & 0.51 \\
        \verb|LESA| & 0.63 & & 0.72 & 0.54 & 0.78 & 0.41 \\
        \verb|NewsClaims| & 0.60 & & 0.77 & 0.42 & 0.80 & 0.36 \\
        \verb|MultiClaim| & 0.52 & & 0.09 & 0.94 & 0.17 & 0.34 \\
        \verb|EnvClaims| & 0.51 & & 0.98 & 0.03 & 0.88 & 0.05 \\
        \verb|GeneralClaim| & 0.68 & & 0.52 & 0.83 & 0.66 & 0.45 \\
        \hline
        \multicolumn{6}{c}{\textbf{CLEF'23}} \\
        \hline
        \verb|CLEF'22| & 0.60 & & 0.99 & 0.20 & 0.77 & 0.32 \\
        \verb|CLEF'23| & 0.86 & & 0.98 & 0.73 & 0.90 & 0.83 \\
        \verb|LESA| & 0.51 & & 0.93 & 0.08 & 0.77 & 0.14 \\
        \verb|NewsClaims| & 0.51 & & 0.94 & 0.08 & 0.78 & 0.13 \\
        \verb|MultiClaim| & 0.72 & & 0.70 & 0.73 & 0.76 & 0.63 \\
        \verb|EnvClaims| & 0.50 & & 1.00 & 0.00 & 0.79 & 0.01 \\
        \verb|GeneralClaim| & 0.82 & & 0.95 & 0.69 & 0.90 & 0.78 \\
        \hline
        \multicolumn{6}{c}{\textbf{GeneralClaim}} \\
        \hline
        \verb|CLEF'22| & 0.55 & & 0.95 & 0.15 & 0.67 & 0.25 \\
        \verb|CLEF'23| & 0.64 & & 0.87 & 0.41 & 0.71 & 0.53 \\
        \verb|LESA| & 0.67 & & 0.88 & 0.45 & 0.71 & 0.58 \\
        \verb|NewsClaims| & 0.60 & & 0.83 & 0.37 & 0.68 & 0.48 \\
        \verb|MultiClaim| & 0.75 & & 0.75 & 0.74 & 0.75 & 0.75 \\
        \verb|EnvClaims| & 0.50 & & 0.98 & 0.01 & 0.66 & 0.02 \\
        \verb|GeneralClaim| & 0.90 & & 0.93 & 0.86 & 0.89 & 0.89 \\
        \hline
        \multicolumn{6}{c}{\textbf{ClaimsBuster}} \\
        \hline
        \verb|CLEF'22| & 0.57 & & 0.99 & 0.14 & 0.90 & 0.24 \\
        \verb|CLEF'23| & 0.91 & & 0.95 & 0.86 & 0.96 & 0.84 \\
        \verb|LESA| & 0.47 & & 0.91 & 0.03 & 0.86 & 0.04 \\
        \verb|NewsClaims| & 0.54 & & 0.95 & 0.12 & 0.89 & 0.18 \\
        \verb|MultiClaim| & 0.69 & & 0.61 & 0.76 & 0.74 & 0.43 \\
        \verb|EnvClaims| & 0.51 & & 1.00 & 0.01 & 0.90 & 0.02 \\
        \verb|GeneralClaim| & 0.86 & & 0.90 & 0.81 & 0.92 & 0.73 \\
        \hline
    \end{tabular}
    \caption{Out-of-distribution evaluation of the \textbf{mDeBERTa} model trained on \textbf{multilingual} configuration comprising of all the 10 selected languages from all tested datasets.}
    \label{tab:out-mdb-all}
\end{table}

\begin{table}
    \centering
    \small
    \setlength\tabcolsep{3pt}
    \begin{tabular}{@{}ccccccc}
        \hline
        \textbf{Train Dataset} & \multicolumn{2}{c}{\textbf{Accuracy}} & \multicolumn{2}{c}{\textbf{Recall}} & \multicolumn{2}{c}{\textbf{F1-score}} \\
        \hline
        English & & & 0: & 1: & 0: & 1: \\
        \hline
        \multicolumn{7}{c}{\textbf{NewsClaims}} \\
        \hline
        \verb|CLEF'22| & 0.60 & & 0.68 & 0.51 & 0.63 & 0.56 \\
        \verb|CLEF'23| & 0.59 & & 0.58 & 0.59 & 0.58 & 0.59 \\
        \verb|LESA| & 0.67 & & 0.67 & 0.66 & 0.67 & 0.66 \\
        \verb|NewsClaims| & 0.42 & & 0.26 & 0.58 & 0.30 & 0.52 \\
        \verb|MultiClaim| & 0.53 & & 0.44 & 0.62 & 0.46 & 0.59 \\
        \verb|EnvClaims| & 0.50 & & 1.00 & 0.00 & 0.67 & 0.00 \\
        \verb|GeneralClaim| & 0.58 & & 0.62 & 0.54 & 0.58 & 0.58 \\
        \hline
        \multicolumn{7}{c}{\textbf{CLEF'22}} \\
        \hline
        \verb|CLEF'22| & 0.64 & & 0.76 & 0.51 & 0.79 & 0.47 \\
        \verb|CLEF'23| & 0.59 & & 0.44 & 0.74 & 0.57 & 0.45 \\
        \verb|LESA| & 0.56 & & 0.24 & 0.87 & 0.37 & 0.43 \\
        \verb|NewsClaims| & 0.49 & & 0.13 & 0.85 & 0.22 & 0.39 \\
        \verb|MultiClaim| & 0.54 & & 0.30 & 0.77 & 0.43 & 0.41 \\
        \verb|EnvClaims| & 0.50 & & 1.00 & 0.00 & 0.85 & 0.00 \\
        \verb|GeneralClaim| & 0.55 & & 0.63 & 0.46 & 0.69 & 0.37 \\
        \hline
        \multicolumn{7}{c}{\textbf{CLEF'23}} \\
        \hline
        \verb|CLEF'22| & 0.60 & & 0.90 & 0.30 & 0.75 & 0.41 \\
        \verb|CLEF'23| & 0.81 & & 0.98 & 0.63 & 0.87 & 0.75 \\
        \verb|LESA| & 0.50 & & 0.99 & 0.01 & 0.79 & 0.02 \\
        \verb|NewsClaims| & 0.49 & & 0.05 & 0.93 & 0.09 & 0.57 \\
        \verb|MultiClaim| & 0.66 & & 0.55 & 0.77 & 0.66 & 0.58 \\
        \verb|EnvClaims| & 0.50 & & 1.00 & 0.00 & 0.80 & 0.00 \\
        \verb|GeneralClaim| & 0.80 & & 0.95 & 0.65 & 0.86 & 0.76 \\
        \hline
        \multicolumn{7}{c}{\textbf{GeneralClaim}} \\
        \hline
        \verb|CLEF'22| & 0.57 & & 0.87 & 0.27 & 0.66 & 0.39 \\
        \verb|CLEF'23| & 0.63 & & 0.89 & 0.37 & 0.70 & 0.50 \\
        \verb|LESA| & 0.65 & & 0.94 & 0.35 & 0.72 & 0.50 \\
        \verb|NewsClaims| & 0.47 & & 0.08 & 0.85 & 0.13 & 0.64 \\
        \verb|MultiClaim| & 0.75 & & 0.69 & 0.81 & 0.73 & 0.77 \\
        \verb|EnvClaims| & 0.50 & & 0.99 & 0.00 & 0.65 & 0.00 \\
        \verb|GeneralClaim| & 0.86 & & 0.92 & 0.79 & 0.85 & 0.85 \\
        \hline
        \multicolumn{7}{c}{\textbf{ClaimsBuster}} \\
        \hline
        \verb|CLEF'22| & 0.63 & & 0.86 & 0.39 & 0.78 & 0.47 \\
        \verb|CLEF'23| & 0.90 & & 0.93 & 0.86 & 0.93 & 0.86 \\
        \verb|LESA| & 0.50 & & 0.99 & 0.00 & 0.78 & 0.00 \\
        \verb|NewsClaims| & 0.49 & & 0.02 & 0.96 & 0.04 & 0.51 \\
        \verb|MultiClaim| & 0.63 & & 0.36 & 0.89 & 0.51 & 0.59 \\
        \verb|EnvClaims| & 0.50 & & 1.00 & 0.00 & 0.79 & 0.00 \\
        \verb|GeneralClaim| & 0.81 & & 0.95 & 0.67 & 0.89 & 0.77 \\
        \hline
    \end{tabular}
    \caption{Out-of-distribution evaluation of the \textbf{XLM-RoBERTa} model trained on a subset of the \textbf{English} language extracted from each tested dataset.}
    \label{tab:xlm-out-en}
\end{table}

\begin{table}
    \centering
    \small
    \setlength\tabcolsep{3pt}
    \begin{tabular}{@{}cccccccc}
        \hline
        \textbf{Train Dataset} & \multicolumn{2}{c}{\textbf{Accuracy}} & \multicolumn{2}{c}{\textbf{Recall}} & \multicolumn{2}{c}{\textbf{F1-score}} \\
        \hline
        Multilingual &  &  & 0: & 1: & 0: & 1: \\
        \hline
        \multicolumn{7}{c}{\textbf{NewsClaims}} \\
        \hline
        \verb|CLEF'22| & 0.48 & & 0.82 & 0.14 & 0.61 & 0.21 & \\
        \verb|CLEF'23| & 0.48 & & 0.48 & 0.48 & 0.48 & 0.48 & \\
        \verb|LESA| & 0.65 & & 0.63 & 0.66 & 0.64 & 0.65 & \\
        \verb|NewsClaims| & 0.80 & & 0.73 & 0.87 & 0.78 & 0.81 & \\
        \verb|MultiClaim| & 0.53 & & 0.26 & 0.79 & 0.35 & 0.62 & \\
        \verb|EnvClaims| & 0.50 & & 1.00 & 0.00 & 0.67 & 0.00 & \\
        \verb|GeneralClaim| & 0.73 & & 0.77 & 0.69 & 0.72 & 0.73 & \\
        \hline
        \multicolumn{7}{c}{\textbf{CLEF'22}} \\
        \hline
        \verb|CLEF'22| & 0.59 & & 0.99 & 0.19 & 0.90 & 0.32 & \\
        \verb|CLEF'23| & 0.71 & & 0.82 & 0.60 & 0.85 & 0.52 & \\
        \verb|LESA| & 0.62 & & 0.70 & 0.54 & 0.77 & 0.40 & \\
        \verb|NewsClaims| & 0.61 & & 0.80 & 0.42 & 0.82 & 0.38 & \\
        \verb|MultiClaim| & 0.51 & & 0.03 & 0.98 & 0.06 & 0.34 & \\
        \verb|EnvClaims| & 0.50 & & 0.99 & 0.01 & 0.88 & 0.02 & \\
        \verb|GeneralClaim| & 0.60 & & 0.72 & 0.47 & 0.77 & 0.37 & \\
        \hline
        \multicolumn{7}{c}{\textbf{CLEF'23}} \\
        \hline
        \verb|CLEF'22| & 0.52 & & 1.00 & 0.03 & 0.80 & 0.06 & \\
        \verb|CLEF'23| & 0.79 & & 0.97 & 0.60 & 0.86 & 0.73 & \\
        \verb|LESA| & 0.50 & & 0.95 & 0.05 & 0.78 & 0.08 & \\
        \verb|NewsClaims| & 0.53 & & 0.88 & 0.17 & 0.76 & 0.24 & \\
        \verb|MultiClaim| & 0.67 & & 0.42 & 0.91 & 0.58 & 0.60 & \\
        \verb|EnvClaims| & 0.50 & & 1.00 & 0.00 & 0.79 & 0.00 & \\
        \verb|GeneralClaim| & 0.84 & & 0.95 & 0.72 & 0.88 & 0.80 & \\
        \hline
        \multicolumn{7}{c}{\textbf{GeneralClaim}} \\
        \hline
        \verb|CLEF'22| & 0.51 & & 0.99 & 0.02 & 0.67 & 0.05 & \\
        \verb|CLEF'23| & 0.63 & & 0.89 & 0.36 & 0.69 & 0.49 & \\
        \verb|LESA| & 0.65 & & 0.89 & 0.40 & 0.72 & 0.53 & \\
        \verb|NewsClaims| & 0.60 & & 0.81 & 0.38 & 0.67 & 0.49 & \\
        \verb|MultiClaim| & 0.71 & & 0.56 & 0.86 & 0.66 & 0.75 & \\
        \verb|EnvClaims| & 0.50 & & 0.99 & 0.00 & 0.66 & 0.00 & \\
        \verb|GeneralClaim| & 0.89 & & 0.93 & 0.84 & 0.88 & 0.88 & \\
        \hline
        \multicolumn{7}{c}{\textbf{ClaimsBuster}} \\
        \hline
        \verb|CLEF'22| & 0.51 & & 1.00 & 0.01 & 0.90 & 0.01 & \\
        \verb|CLEF'23| & 0.86 & & 0.98 & 0.73 & 0.96 & 0.80 & \\
        \verb|LESA| & 0.50 & & 0.96 & 0.03 & 0.88 & 0.04 & \\
        \verb|NewsClaims| & 0.53 & & 0.89 & 0.17 & 0.86 & 0.20 & \\
        \verb|MultiClaim| & 0.62 & & 0.31 & 0.93 & 0.46 & 0.37 & \\
        \verb|EnvClaims| & 0.51 & & 1.00 & 0.01 & 0.90 & 0.02 & \\
        \verb|GeneralClaim| & 0.83 & & 0.90 & 0.76 & 0.92 & 0.70 & \\
        \hline
    \end{tabular}
    \caption{Out-of-distribution evaluation of the \textbf{XLM-RoBERTa} model trained on \textbf{multilingual} configuration comprising of all the 10 selected languages from all tested datasets.}
    \label{tab:xlm-out-all}
\end{table}

\begin{table}
    \centering
    \small
    \setlength\tabcolsep{3pt}
    \begin{tabular}{@{}ccccccc}
        \hline
        \textbf{Train Dataset} & \multicolumn{2}{c}{\textbf{Accuracy}} & \multicolumn{2}{c}{\textbf{Recall}} & \multicolumn{2}{c}{\textbf{F1-score}} \\
        \hline
        English & & & 0: & 1: & 0: & 1: \\
        \hline
        \multicolumn{5}{c}{\textbf{NewsClaims}} \\
        \hline
        \verb|CLEF'22| & 0.50 & & 1.00 & 0.00 & 0.67 & 0.00 \\
        \verb|CLEF'23| & 0.58 & & 0.40 & 0.76 & 0.49 & 0.65 \\
        \verb|LESA| & 0.63 & & 0.52 & 0.74 & 0.58 & 0.67 \\
        \verb|NewsClaims| & 0.63 & & 0.51 & 0.74 & 0.58 & 0.67 \\
        \verb|MultiClaim| & 0.53 & & 0.55 & 0.51 & 0.51 & 0.54 \\
        \verb|EnvClaims| & 0.52 & & 0.98 & 0.05 & 0.66 & 0.09 \\
        \verb|GeneralClaim| & 0.55 & & 0.23 & 0.87 & 0.33 & 0.69 \\
        \hline
        \multicolumn{5}{c}{\textbf{CLEF'22}} \\
        \hline
        \verb|CLEF'22| & 0.50 & & 1.00 & 0.00 & 0.85 & 0.00 \\
        \verb|CLEF'23| & 0.57 & & 0.45 & 0.69 & 0.57 & 0.43 \\
        \verb|LESA| & 0.59 & & 0.30 & 0.87 & 0.45 & 0.45 \\
        \verb|NewsClaims| & 0.59 & & 0.28 & 0.90 & 0.43 & 0.46 \\
        \verb|MultiClaim| & 0.52 & & 0.09 & 0.95 & 0.16 & 0.42 \\
        \verb|EnvClaims| & 0.55 & & 0.88 & 0.21 & 0.82 & 0.27 \\
        \verb|GeneralClaim| & 0.52 & & 0.07 & 0.97 & 0.13 & 0.42 \\
        \hline
        \multicolumn{5}{c}{\textbf{CLEF'23}} \\
        \hline
        \verb|CLEF'22| & 0.50 & & 1.00 & 0.00 & 0.80 & 0.00 \\
        \verb|CLEF'23| & 0.68 & & 0.97 & 0.39 & 0.85 & 0.54 \\
        \verb|LESA| & 0.52 & & 0.99 & 0.05 & 0.80 & 0.09 \\
        \verb|NewsClaims| & 0.52 & & 0.99 & 0.05 & 0.80 & 0.09 \\
        \verb|MultiClaim| & 0.65 & & 0.65 & 0.65 & 0.71 & 0.56 \\
        \verb|EnvClaims| & 0.50 & & 1.00 & 0.00 & 0.80 & 0.00 \\
        \verb|GeneralClaim| & 0.74 & & 0.86 & 0.61 & 0.84 & 0.65 \\
        \hline
        \multicolumn{5}{c}{\textbf{GeneralClaim}} \\
        \hline
        \verb|CLEF'22| & 0.50 & & 1.00 & 0.00 & 0.66 & 0.00 \\
        \verb|CLEF'23| & 0.69 & & 0.82 & 0.56 & 0.72 & 0.65 \\
        \verb|LESA| & 0.65 & & 0.94 & 0.36 & 0.72 & 0.50 \\
        \verb|NewsClaims| & 0.66 & & 0.94 & 0.38 & 0.73 & 0.53 \\
        \verb|MultiClaim| & 0.74 & & 0.79 & 0.68 & 0.73 & 0.73 \\
        \verb|EnvClaims| & 0.55 & & 1.00 & 0.09 & 0.65 & 0.16 \\
        \verb|GeneralClaim| & 0.79 & & 0.77 & 0.80 & 0.78 & 0.79 \\
        \hline
        \multicolumn{5}{c}{\textbf{ClaimsBuster}} \\
        \hline
        \verb|CLEF'22| & 0.50 & & 1.00 & 0.00 & 0.79 & 0.00 \\
        \verb|CLEF'23| & 0.80 & & 0.86 & 0.74 & 0.86 & 0.74 \\
        \verb|LESA| & 0.57 & & 1.00 & 0.13 & 0.81 & 0.23 \\
        \verb|NewsClaims| & 0.57 & & 1.00 & 0.13 & 0.81 & 0.23 \\
        \verb|MultiClaim| & 0.70 & & 0.83 & 0.57 & 0.80 & 0.60 \\
        \verb|EnvClaims| & 0.51 & & 1.00 & 0.02 & 0.79 & 0.04 \\
        \verb|GeneralClaim| & 0.78 & & 0.71 & 0.85 & 0.79 & 0.72 \\
        \hline
    \end{tabular}
    \caption{Out-of-distribution evaluation of the \textbf{LESA} model trained on a subset of the \textbf{English} language.}
    \label{tab:out-lesa-en  }
\end{table}

\begin{table}
    \centering
    \small
    \setlength\tabcolsep{3pt}
    \begin{tabular}{@{}cccccccc}
        \hline
        \textbf{Train Dataset} & \multicolumn{2}{c}{\textbf{Accuracy}} & \multicolumn{2}{c}{\textbf{Recall}} & \multicolumn{2}{c}{\textbf{F1-score}} \\
        \hline
        Multilingual &  &  & 0: & 1: & 0: & 1: \\
        \hline
        \multicolumn{6}{c}{\textbf{NewsClaims}} \\
        \hline
        \verb|CLEF'22| &  0.51 & & 1.00 & 0.01 & 0.67 & 0.02 \\
        \verb|CLEF'23| & 0.50 & & 0.35 & 0.65 & 0.41 & 0.56 \\
        \verb|LESA| & 0.58 & & 0.44 & 0.72 & 0.51 & 0.63 \\
        \verb|NewsClaims| & 0.62 & & 0.39 & 0.85 & 0.51 & 0.69 \\
        \verb|MultiClaim| & 0.52 & & 0.42 & 0.62 & 0.46 & 0.56 \\
        \verb|EnvClaims| & 0.50 & & 0.98 & 0.01 & 0.66 & 0.03 \\
        \verb|GeneralClaim| & 0.55 & & 0.36 & 0.73 & 0.44 & 0.61 \\
        \hline
        \multicolumn{6}{c}{\textbf{CLEF'22}} \\
        \hline
        \verb|CLEF'22| & 0.52 & & 0.98 & 0.05 & 0.88 & 0.09 \\
        \verb|CLEF'23| & 0.61 & & 0.54 & 0.68 & 0.67 & 0.39 \\
        \verb|LESA| & 0.54 & & 0.17 & 0.90 & 0.28 & 0.35 \\
        \verb|NewsClaims| & 0.58 & & 0.26 & 0.89 & 0.40 & 0.37 \\
        \verb|MultiClaim| & 0.51 & & 0.05 & 0.96 & 0.10 & 0.34 \\
        \verb|EnvClaims| & 0.50 & & 0.99 & 0.01 & 0.88 & 0.02 \\
        \verb|GeneralClaim| & 0.52 & & 0.09 & 0.95 & 0.17 & 0.34 \\
        \hline
        \multicolumn{6}{c}{\textbf{CLEF'23}} \\
        \hline
        \verb|CLEF'22| & 0.50 & & 1.00 & 0.00 & 0.80 & 0.00 \\
        \verb|CLEF'23| & 0.67 & & 0.97 & 0.37 & 0.85 & 0.52 \\
        \verb|LESA| & 0.58 & & 0.98 & 0.17 & 0.76 & 0.28 \\
        \verb|NewsClaims| & 0.59 & & 0.89 & 0.28 & 0.79 & 0.38 \\
        \verb|MultiClaim| & 0.68 & & 0.73 & 0.63 & 0.73 & 0.63 \\
        \verb|EnvClaims| & 0.50 & & 1.00 & 0.00 & 0.80 & 0.00 \\
        \verb|GeneralClaim| & 0.70 & & 0.88 & 0.52 & 0.79 & 0.62 \\
        \hline
        \multicolumn{6}{c}{\textbf{GeneralClaim}} \\
        \hline
        \verb|CLEF'22| & 0.50 & & 1.00 & 0.00 & 0.67 & 0.00 \\
        \verb|CLEF'23| & 0.66 & & 0.84 & 0.48 & 0.71 & 0.58 \\
        \verb|LESA| & 0.65 & & 0.91 & 0.39 & 0.72 & 0.53 \\
        \verb|NewsClaims| & 0.64 & & 0.74 & 0.53 & 0.66 & 0.60 \\
        \verb|MultiClaim| & 0.72 & & 0.72 & 0.72 & 0.72 & 0.72 \\
        \verb|EnvClaims| & 0.50 & & 0.99 & 0.00 & 0.66 & 0.01 \\
        \verb|GeneralClaim| & 0.78 & & 0.85 & 0.71 & 0.79 & 0.76 \\
        \hline
        \multicolumn{6}{c}{\textbf{ClaimsBuster}} \\
        \hline
        \verb|CLEF'22| & 0.50 & & 1.00 & 0.00 & 0.90 & 0.00 \\
        \verb|CLEF'23| & 0.74 & & 0.89 & 0.58 & 0.90 & 0.56 \\
        \verb|LESA| & 0.58 & & 0.90 & 0.26 & 0.86 & 0.32 \\
        \verb|NewsClaims| & 0.64 & & 0.69 & 0.59 & 0.78 & 0.40 \\
        \verb|MultiClaim| & 0.62 & & 0.56 & 0.68 & 0.69 & 0.40 \\
        \verb|EnvClaims| & 0.51 & & 1.00 & 0.01 & 0.89 & 0.02 \\
        \verb|GeneralClaim| & 0.70 & & 0.78 & 0.61 & 0.83 & 0.46 \\
        \hline
    \end{tabular}
    \caption{Out-of-distribution evaluation of the \textbf{LESA} model trained on \textbf{multilingual} configuration comprising of all the 10 selected languages from all tested datasets.}
    \label{tab:out-lesa-all}
\end{table}

\end{document}